\documentclass[letterpaper]{article} 
\usepackage[]{arxiv}  
\usepackage{times}  
\usepackage{helvet}  
\usepackage{courier}  
\usepackage[hyphens]{url}  
\usepackage{graphicx} 
\usepackage{natbib}  
\usepackage{caption} 
\usepackage{algorithm}
\usepackage{algorithmic}
\usepackage{amssymb} 
\usepackage{amsmath} 
\usepackage{tikz}
\usepackage{pgfplots} 
\usepackage{multirow} 

\usepackage{newfloat}
\usepackage{listings}
\DeclareCaptionStyle{ruled}{labelfont=normalfont,labelsep=colon,strut=off} 
\floatstyle{ruled}
\newfloat{listing}{tb}{lst}{}
\floatname{listing}{Listing}

\title{Learning Regularizers: Learning Optimizers that can Regularize}
\author{
    Suraj K Sahoo \textsuperscript{\rm 1}
    Narayanan C. Krishnan \textsuperscript{\rm 1}
}
\affiliations{
    \textsuperscript{\rm 1} Mehta Family School of Data Science and Artificial Intelligence, Department of Data Science, Indian Institute of Technology Palakkad, Kerala, India\\

}

\begin{document}

\maketitle

\begin{abstract}
Learned Optimizers (LOs), a type of Meta-learning, have gained traction due to their ability to be parameterized and trained for efficient optimization. Traditional gradient-based methods incorporate explicit regularization techniques such as Sharpness-Aware Minimization (SAM), Gradient-norm Aware Minimization (GAM), and Gap-guided Sharpness-Aware Minimization (GSAM) to enhance generalization and convergence. In this work, we explore a fundamental question: \textbf{Can regularizers be learned?} We empirically demonstrate that LOs can be trained to learn and internalize the effects of traditional regularization techniques without explicitly applying them to the objective function. We validate this through extensive experiments on standard benchmarks (including MNIST, FMNIST, CIFAR and Neural Networks such as MLP, MLP-Relu and CNN), comparing LOs trained with and without access to explicit regularizers. Regularized LOs consistently outperform their unregularized counterparts in terms of test accuracy and generalization. Furthermore, we show that LOs retain and transfer these regularization effects to new optimization tasks by inherently seeking minima similar to those targeted by these regularizers. Our results suggest that LOs can inherently learn regularization properties, \textit{challenging the conventional necessity of explicit optimizee loss regularization.}
\end{abstract}


\section{Introduction}
Stochastic Gradient Descent (SGD) and its variants, while effective for neural network training, often require regularization for improved convergence and generalization. Recent geometry-aware regularizers such as Sharpness-Aware Minimization (SAM) \cite{sam}, Gradient Norm Aware Minimization (GAM) \cite{gam}, and Surrogate Gap Guided Sharpness-Aware Minimization (GSAM) \cite{gsam} promote smoother minima and mitigate overfitting. However, these regularizers introduce manual tuning overhead and additional computational complexity.

An emerging alternative to hand-crafted SGD is Learned Optimizers (LOs), which aim to discover effective update rules. Early works, such as \cite{L2LGDGD}, introduced LSTM-based architectures to capture gradient history for more informed updates. Over time, architectures like multi-layer perceptrons (MLPs) \cite{mlpLO} and hybrid models combining RNNs with MLPs due to \cite{hierarchyRNN}, \cite{hierarchyMLP}, techniques such as Curriculum and Imitation Learning techniques \cite{Curriculum_chen2020} have been explored to enhance LO capabilities. Beyond architectural innovations, LOs have been observed to exhibit interpretable behaviors such as momentum-like dynamics and adaptive learning rate schedules, akin to traditional optimization methods \cite{maheswaranathan2021reverse}. Nevertheless, LO training can be unstable. Stability regularization techniques by minimizing the effect of small perturbations, improved robustness \cite{smoothDM}.

Despite the promise of LOs, a crucial question remains regarding their generalizability and the role of regularization in the optimization process. While prior work has demonstrated that optimizers can be learned, we extend this inquiry by asking: \textbf{Can regularization be learned?} If an LO can inherently develop regularization capabilities, it could enable a more adaptive and automated approach to optimization, reducing the reliance on explicit regularization techniques while still achieving stability and generalization.

Building on this idea, \textit{we empirically demonstrate that LOs can inherently learn regularization principles}. Specifically, we show that an LO trained with a regularization objective can \textit{internalize and transfer} this property to downstream tasks, enforcing regularization at the optima without explicit manual intervention. By embedding regularization directly into the optimization process, LOs eliminate the overhead of hand-crafted regularizers, offering a more scalable and efficient approach to training deep models.

\section{Preliminaries}

\subsection{Learning Optimizers (LOs)}  \label{section:lo}
Learning a Learned Optimizer (LO) requires a hierarchical learning process in which the optimizer itself is refined through experience. This process consists of two phases: the \textbf{Meta-Training Phase}, where the LO is trained, and the \textbf{Meta-Test Phase}, where the LO is evaluated on downstream learning tasks \cite{bilevel}. The \textbf{meta-training phase} involves two key steps: \textit{inner training} and \textit{outer training}. \textbf{Inner training} refers to the optimization of the \textit{optimizee}, which serves as a data point for training the LO during the outer training. The optimizee operates on a supervised learning dataset consisting of input-label pairs \(\{(x_i, y_i)\}_{i=1}^{N}\), where \(x_i \in \mathbb{R}^d\) is an input feature vector, \(y_i\) is the corresponding label, and \(N\) is the number of training samples. The optimizee's objective is to minimize a loss function \(\mathcal{L}(\theta)\), defined as:
    \begin{equation}\label{eqn:LO}
    \mathcal{L}(\theta) = \frac{1}{N} \sum_{i=1}^{N} \ell(g(x_i; \theta), y_i)
    \end{equation}
    where \(g(x_i;\theta)\) represents the model’s predicted probability distribution over classes, and \(\ell(\cdot, \cdot)\) denotes the loss function (e.g., cross entropy for classification).
\textbf{Outer training} optimizes the LO itself by learning to generate parameter updates that lead to efficient convergence of the optimizee.

We adopt an RNN-based LO, $F$ parameterized by $\phi$, to train an optimizee for a classification task. At each inner optimization step, the optimizee's parameters (\(\theta\)) are updated using the LO’s output. The LO takes as input the gradient of the optimizee’s loss function (\(\nabla_{\theta_t} \mathcal{L}(\theta_t)\)), the current parameters (\(\theta_t\)), and the hidden state of the RNN from the previous iteration (\(h_{t-1}\)) to update $\theta$:
\begin{equation}
\theta_{t+1} = \theta_t + F \left( \nabla_{\theta_t} \mathcal{L}(\theta_t), \theta_t, h_{t-1} ; \phi \right)
\end{equation}
The hidden state \(h_{t-1}\) allows the LO to retain historical information, enabling more sophisticated update strategies than traditional optimizers. To ensure scalability across optimizees with different parameter dimensions, the LO operates in a \textit{coordinate-wise} manner, meaning the same RNN is applied independently to each parameter coordinate. This design allows the LO to generalize across different models and tasks.

Outer training involves learning \(\phi\) to minimize the optimizee’s loss function \(\mathcal{L}(\theta)\) efficiently across training steps. The objective is to minimize the cumulative loss over the trajectory of optimizee updates:
\begin{equation}
\mathcal{L}_{\text{meta}}(\phi) = \sum_{t=1}^{T} w_t \mathcal{L}(\theta_t)
\end{equation}
where \(w_t\) are weighting coefficients that can be uniform or decayed to emphasize performance at later steps. The meta-gradient for updating \(\phi\) is computed using \textit{truncated backpropagation through time (BPTT)}, as differentiating through long optimization trajectories is computationally prohibitive. The update rule for the LO parameters is:
\begin{equation}
\phi \leftarrow \phi - \eta \nabla_{\phi} \mathcal{L}_{\text{meta}}(\phi)
\end{equation}
where \(\eta\) is the meta-learning rate.

Once trained, the LO is evaluated in the \textbf{meta-test phase}, where it is used to optimize an \textit{unseen} optimizee for a downstream task. The LO's effectiveness is measured by how well it generalizes to new datasets and architectures beyond those seen during meta-training. \textbf{In this phase, the optimizee is trained on a new dataset of input-label pairs \(\{(x_j, y_j)\}_{j=1}^{M}\), where \(M\) is the number of test samples}. The LO is expected to accelerate convergence and improve generalization.

\subsection{Regularization}\label{sec:regularization}
Regularization enhances model generalization by mitigating overfitting through constraints or modifications during training. Traditional methods include  $L_1$ \cite{l1_reg_tibshirani}  and  $L_2$ \cite{L2_reg_ridge} weight penalties, data augmentation \cite{data_augmentation}, dropout\cite{dropout}, and normalization techniques like batch and layer normalization \cite{batch_norm} \cite{layer_normalization}. Recent work explores the connection between loss landscape geometry and generalization \cite{asym_valley, 2018visualizinglosslandscape}, leading to methods SAM, GSAM, GAM and many more. 

SAM explicitly promotes flatter minima by optimizing for the worst-case loss within a small neighborhood. Formally, let $\mathcal{L}: \Theta \times X \times Y \rightarrow \mathbb{R}$ be the loss function defined over the parameter space $\Theta$, input space $X$, and output space $Y$. SAM modifies the empirical loss (\ref{eqn:LO}) to:

\begin{equation}
    \hat{\mathcal{L}}^{\text{SAM}}(\theta) = \max_{\|\epsilon\| \leq \rho} \hat{\mathcal{L}}(\theta + \epsilon),
\end{equation}

where $\rho$ defines the neighborhood radius. This approach seeks parameters $\theta$ that reside in regions of the loss landscape where the loss remains low within the specified neighborhood, thereby avoiding sharp minima that could lead to poor generalization. 

Surrogate Gap Guided Sharpness-Aware Minimization (GSAM) introduces a surrogate gap to address limitations stemming from the fact that SAM’s perturbed-loss objective treats sharp and flat minima similarly, where both can yield equally low regularized loss values, so SAM alone cannot reliably favor flat minima. The surrogate gap,
\[
h(\theta) = \hat{\mathcal{L}}^{\text{SAM}}(\theta) - \hat{\mathcal{L}}(\theta).
\]
enables direct minimization of sharpness without affecting the worst-case perturbed loss, by decomposing the gradient into components that are parallel and orthogonal to the perturbation direction and performing an orthogonal ascent to reduce this gap. Calculating the gradient involves two steps, including a gradient descent as is necessary in the case of SAM and another step to find the orthogonal direction to minimize the surrogate gap without hampering $\mathcal{L}^{SAM}$. 

Gradient Norm Aware Minimization (GAM) focuses on first-order information of the loss landscape by penalizing the maximum norm of the loss gradient within a neighborhood. This approach aims to control the sharpness of the loss surface by ensuring that the gradient norm does not escalate drastically in the vicinity of the current parameters. The regularized loss is defined as:

\[
\hat{\mathcal{L}}^{\text{GAM}}(\theta) = \hat{\mathcal{L}}(\theta) + \max_{\|\epsilon\| \leq \rho} \|\nabla \hat{\mathcal{L}}(\theta + \epsilon)\|,
\]

where $\rho$ specifies the neighborhood radius.

These methods, while powerful, come with additional computational overhead due to the necessity of calculating auxiliary objectives or penalties at each iteration. For instance, GSAM requires computing the worst-case perturbation in a predefined neighborhood, which involves additional forward and backward passes: one for generating the adversarial perturbation and a second for computing the perturbed loss gradient and then projects and decomposes the original gradient to apply an additional orthogonal ascent step to minimize the surrogate gap (without affecting the perturbed loss). Despite these computational demands, the benefits of such methods in guiding the optimization process toward smoother minima have made them an integral part of modern optimization strategies. \textit{As a result, exploration as ours in this work on whether these regularization strategies can be internalized by Learned Optimizers (LOs), would eliminate the need for explicit regularizer computations and reducing the overall computational burden.}

\textbf{Note:} Generalizability in optimization has two key aspects, both crucial for LOs. The first, \textit{optimizer generalization}, refers to an LO’s adaptability across diverse optimizee tasks \cite{aistat2023, optimizerGen}, ensuring effectiveness across different problem domains. The second, \textit{optimizee generalization}, the focus of this work, involves guiding the optimizee toward parameter configurations that generalize well to unseen data \cite{aistat2023}. In this work, we \textit{explore whether LOs can implicitly learn regularization, improving optimizee generalization without explicitly incorporating regularization in the meta-test phase.}

\section{Learning a Regularizer}\label{section:l2g}

The above mentioned regularizers SAM, GSAM or GAM, add a penalty term to the loss function, referred to as the regularized loss, which discourages convergence to sharp minima. The penalty increases the total loss at sharp minima, making them less favorable compared to flatter ones. This modification, in turn, guides the optimizer toward flatter regions, which are associated with better generalization.

\begin{figure}[!htb]
    \centering
    \includegraphics[width=\linewidth]{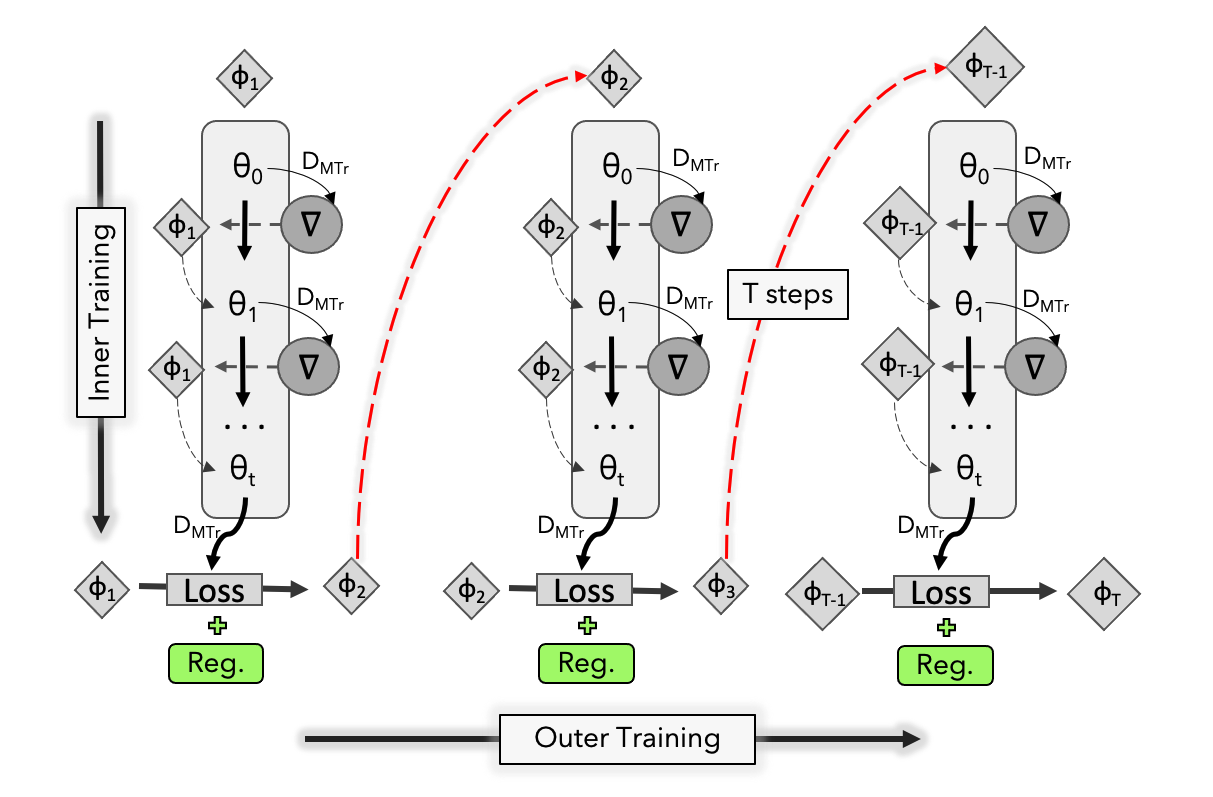}
    \caption{Learning a Regularizer}
    \label{fig:LO-training}
\end{figure}

Inspired by this approach, we extend the Learned Optimizer framework by incorporating a similar regularization term into LO's loss, encouraging it to favor flatter solutions during training. We modify the optimizer’s loss function by adding a sharpness-aware regularization term \( \mathcal{L}_{\text{reg}}(\phi) \in \{ \mathcal{L}_{SAM}, \mathcal{L}_{GSAM}, \mathcal{L}_{GAM} \} \), which captures the structural property enforced by the regularizer. The penalty function $\mathcal{L}_{\text{reg}}(\phi)$ is derived based on the desired property of the optimizee's loss landscape, as discussed in previous section. The optimizer is thus trained to minimize not just the empirical loss but also the regularization penalty, shaping its hypothetical update rule accordingly. Figure \ref{fig:LO-training} illustrates the step in which the regularization term is added, depicting the overall structure of the training process.




\textbf{Smoothing Regularization}: To ensure stability in optimizer updates, we adopt a perturbation-based regularization strategy \cite{smoothDM} inspired by adversarial training . Given the optimizee state (the input to the optimizer, in our case, gradient) at time step \( t \), denoted as \( s_t \), we construct a perturbed state \( s'_t \) such that \( s'_t \in B(s_t, \epsilon) \), where \( B(s_t, \epsilon) \) is the set of states within an \( \epsilon \)-radius under the \( \ell_{\infty} \) norm. Fixing the hidden state \( h_t \), the corresponding parameter increments \( u_t \) and \( u'_t \) for \( s_t \) and \( s'_t \) can be written explicitly as functions of the state, i.e., \( u_t = u(s_t) \) and \( u'_t = u(s'_t) \). To promote smoothness and robustness, we directly minimize the worst-case discrepancy \( \max d(u(s_t), u(s'_t)) \), encouraging the optimizer to produce similar updates for neighboring states. The smooth regularization loss can therefore be defined as:    
\begin{equation}
    \mathcal{L}_{\text{smooth}}(\phi) = \max_{s'_t \in B(s_t, \epsilon)} \| u(s_t) - u(s'_t) \|^2.
\end{equation}

To efficiently compute the worst-case discrepancy \( \mathcal{L}_{\text{smooth}}(\phi) \), we employ an iterative projected gradient ascent (PGA). Specifically, we iteratively update the perturbed state \( s'_t \) within the bounded \( \epsilon \)-ball using the following procedure: $ s'_t \leftarrow s'_t + \alpha \cdot \text{sign}(\nabla_{s'_t} \mathcal{L}_{\text{smooth}}(\phi))$ where \( \alpha \) is the step size. To ensure that \( s'_t \) remains within the allowable perturbation bound, we apply the projection step: $ s'_t = \text{Proj}_{B(s_t, \epsilon)}(s'_t)$, where projection ensures that \( s'_t \) stays within the \( \epsilon \)-radius of \( s_t \). This iterative procedure is executed for a fixed number of steps, \( N_{\text{PGA}} \), refining the perturbation towards the worst-case deviation. This ensures that the optimizer produces stable parameter updates, reducing sensitivity to small variations in the optimizee’s state and improving generalization. 

The resulting meta-regularized objective now becomes:
\begin{equation}
\begin{aligned}
\phi^* &= \arg\min_{\phi} \mathcal{L}_{\text{meta, reg}}(\phi), \text{  where},\\
\mathcal{L}_{\text{meta, reg}}(\phi) &= \mathcal{L}_{\text{meta}}(\phi) 
+ \lambda_{\text{smooth}} \mathcal{L}_{\text{smooth}}(\phi) 
+ \lambda_{\text{reg}} \mathcal{L}_{\text{reg}}(\phi)
\end{aligned}
\end{equation}
where $\lambda_{\text{smooth}}$ and $\lambda_{\text{reg}}$ are the hyperparameters associated with the smoothing and flatness aware regularization objectives respectively.

\section{Early Evidence}
\begin{figure}[htbp]
    \centering
    \includegraphics[width=\linewidth]{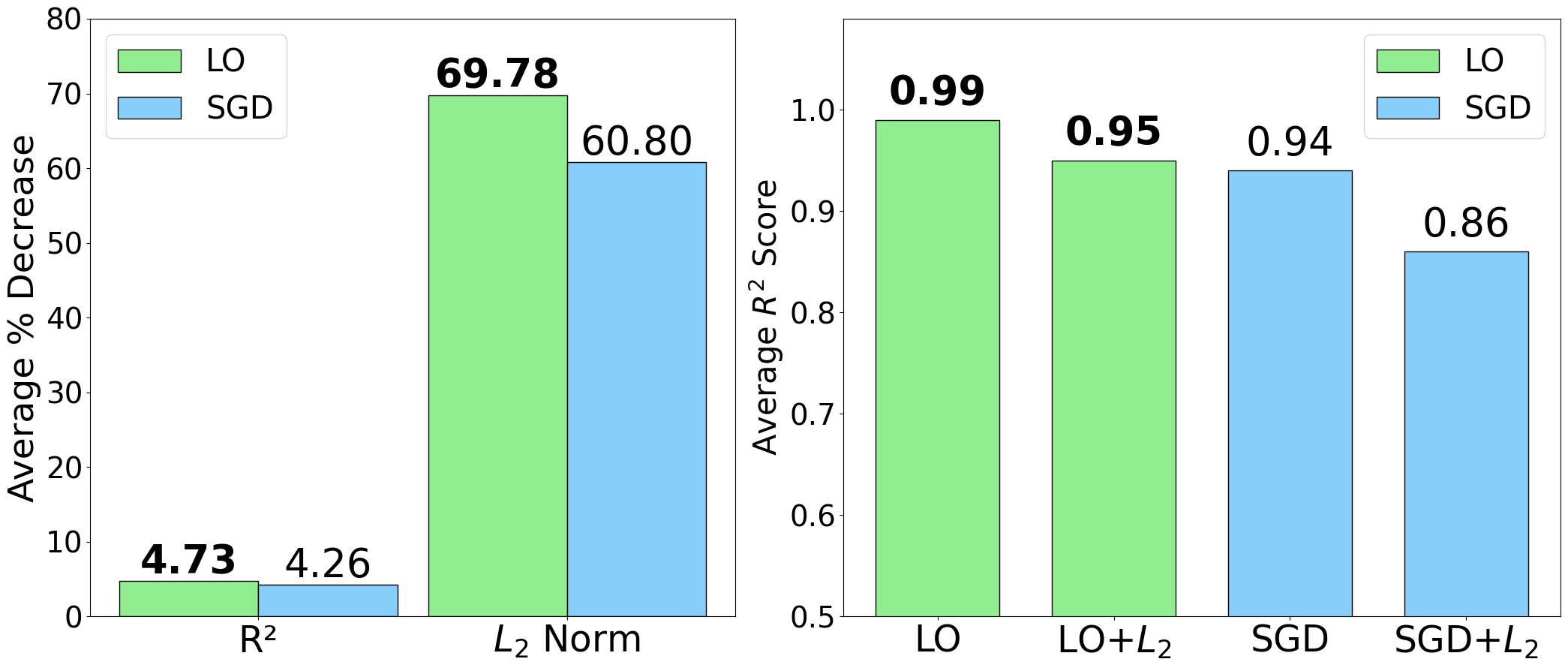}
    \caption{Effect of $L_2$ regularization on LO and SGD. $L_2$ regularization significantly reduces the norm of model parameters. The left panel shows the average of percentage decrease in the parameter norm and corresponding $R^2$ score across each regression task. The right panel compares $R^2$ scores for LO and SGD with and without $L_2$ regularization.}
    \label{fig:l2_reg_comparison}
\end{figure}

We carried out preliminary experiments to evaluate our hypothesis, specifically with $L_2$ regularization using the base optimizer as SGD. Data points were generated from multiple noisy third-degree polynomials for a regression task. The models obtained using both with and without regularization were evaluated on the same unseen test dataset for fair comparison. As expected, Figure \ref{fig:l2_reg_comparison} shows that applying $L_2$ regularization significantly reduces the $L_2$-norm compared to the unregularized case.

Similarly, an LO (LSTM-based architecture as introduced in \cite{L2LGDGD}) was meta-trained on a set of regression problems (curve fitting), with datasets generated  from random third degree polynomials. The LO was meta-tested on a similar randomly generated test dataset using third-degree polynomials. As shown in Figure \ref{fig:l2_reg_comparison}, the curve fitted using a LO trained without $L_2$ regularization converges to parameters with higher magnitudes compared to a LO trained with regularized loss, \textbf{even though the regularization was not applied during meta-test}.

These results provide compelling early evidence of an LO's ability to adapt its parameters to incorporate additional regularization. The detailed setting for this set of experiments (for both SGD and LO), along with the visualization of the learned fits, are provided in the supplementary material. 

\section{Experiments and Results}
In this section, we outline the experimental setup used to test the hypothesis that regularization (such as SAMs, GAM, or GSAM) can be learned. To enhance training stability and extend optimization across multiple tasks, we employed \textbf{curriculum learning} \cite{Curriculum_chen2020}. This method mitigates overfitting on the unrolling length ($T$), which is usually set to 20, by dynamically adjusting the number of unrolling steps during training and evaluation.

For consistency, we trained the LO on the same task across all regularizers: \textbf{MNIST classification using a Multi-Layer Perceptron (MLP)} with a single hidden layer of 20 neurons and a sigmoid activation function. The dataset was split into three parts: 50\% of the training data for the \textbf{Meta-Training} phase, the remaining 50\% of the training data for \textbf{Meta-Test Training}, and the test set for \textbf{Meta-Test Testing}. The performance of the trained LO is evaluated in the \textbf{Meta-Testing Phase}. This phase consists of two steps: Meta-Test Training and Meta-Test Testing. In the former, the trained LO is employed to optimize various tasks (optimizees). The data used for this phase are the Meta-Test Train Data. In \textbf{Meta-Test Testing}, the optimized tasks are evaluated on the Meta-Test Test Data to assess the generalization capabilities of the LO.

Due to high computational and memory requirements (high GPU memory) while working with LOs we restrict the experiments to datasets and architectures used for evaluating LOs to make it manageable for 3 different neural network architectures with 3 different tasks. The tasks selected for evaluation include MNIST, Fashion MNIST (FMNIST) and CIFAR-10 Classification, whose computational and memory requirements are comparable to MNIST, which was used for training the LO. We used an MLP (single layer of 20 neurons) with sigmoid activation, an MLP with ReLU activation, and a Convolutional Neural Network (CNN) comprising two 2D convolutional layers with 16 $3\times3$ kernels and 32 $5\times5$ kernels, respectively, with a 2D max-pooling layer (kernel size $2\times2$) in between. Each task underwent 10 independent runs to ensure robust evaluation. The classification accuracies obtained for the different datasets, across different architectures, by the LO with/without the regularizer are presented in Table \ref{tab:accuracy_results}. We can clearly observe that the performance of the optimizee model is better when the LO with the regularizer is used to train the optimizee model.

\subsection{Evaluation of Regularization Properties at Points of Convergence}


Measuring the regularization properties for chosen regularizers at point of convergence requires a search in the vicinity of the optima. We first applied a 10-step Projected Gradient Ascent (PGA) to analyze changes in the loss or gradient norm at perturbed points relative to the point of convergence (PoC). Specifically, for SAM and GSAM, we employ PGA to iteratively perturb the parameters of a model within a defined neighborhood to maximize the validation loss. The magnitude of the perturbation is limited by $\epsilon$ = 0.1, with each update step set to $\frac{\epsilon}{10}$. Perturbations are applied under the $L_{\infty}$ norm, limiting the maximum change per parameter. Initially, parameters are perturbed with random Gaussian noise to explore the loss landscape, followed by 10 PGA iterations to approach the configuration yielding the highest validation loss within the specified constraints. For GAM, we assessed the difference in gradient norm at a perturbed point from the PoC. We examined the effect of varying the neighborhood size around the PoC by considering different ball radii: \([0.001, 0.005, 0.01, 0.05, 0.1]\), while estimating the maximum value of loss or gradient norm. Furthermore, We conducted a temporal analysis by measuring these properties at two key stages of training:
\begin{itemize}
    \item At convergence, defined by a stopping criterion where the optimizee's training loss does not consistently decrease for 100 consecutive steps.
    \item After a fixed number of training iterations, regardless of convergence status.
\end{itemize}  

\begin{figure}[htbp]
    \centering
    \includegraphics[width=0.5\textwidth]{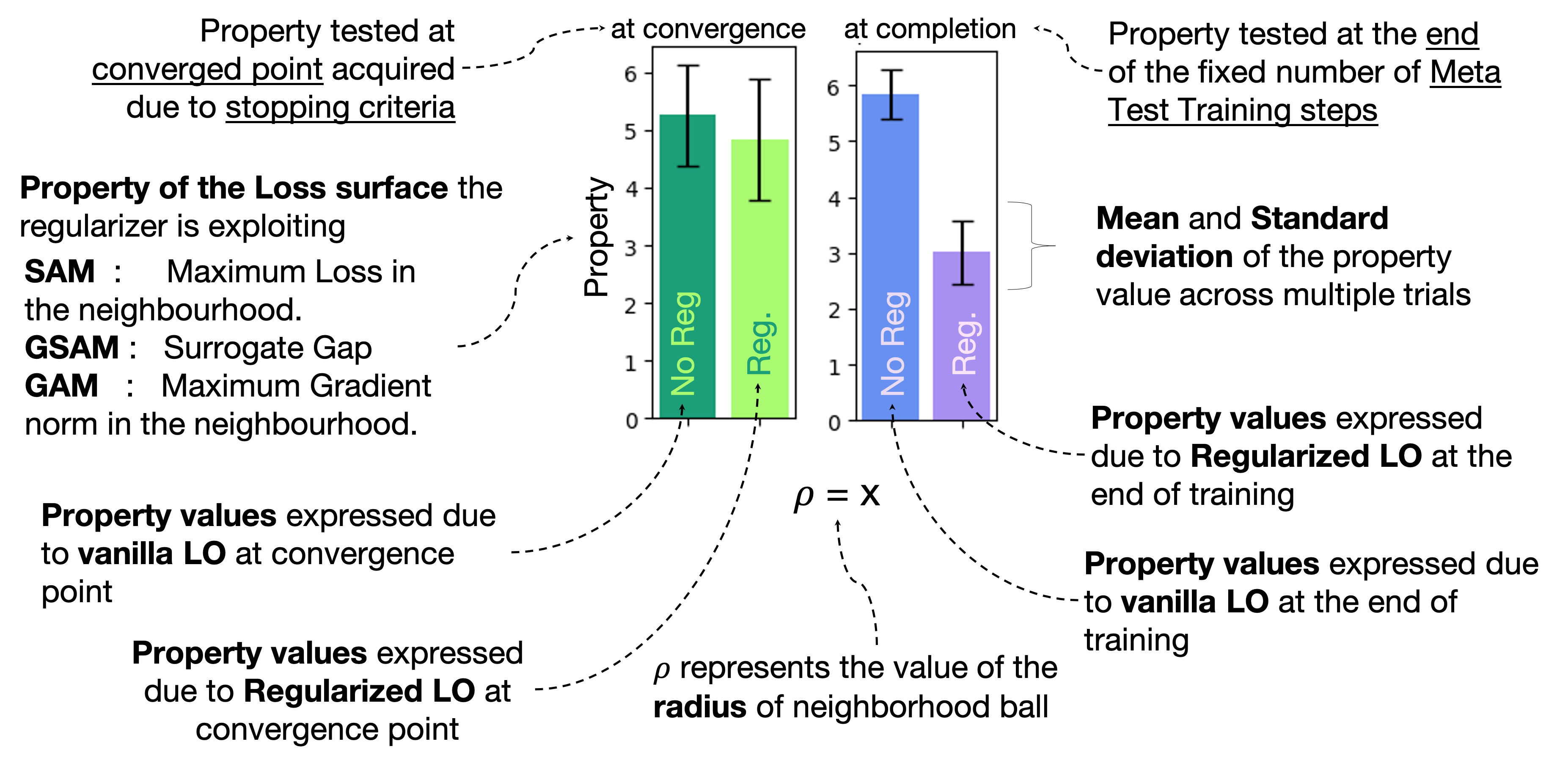}
    \caption{Labeled diagram for understanding of the plots}
    \label{fig:result_interpret}
\end{figure}

We use the illustration in Figure \ref{fig:result_interpret} as the template for presenting the results on the regularization properties. Our analysis focuses on the `property satisfaction' metrics across various tasks, with particular emphasis on the results obtained at the conclusion of the projected gradient ascent steps. This approach facilitates a more intuitive understanding of the influence of each regularizer, especially when examined from the perspective of the loss landscape.

\begin{table}[t]
\centering
\resizebox{\columnwidth}{!}{%
\begin{tabular}{|c|c|l|l|}
\hline
\textbf{Dataset} & \textbf{Architecture} & \textbf{Regularizer} & \textbf{Accuracy} \\
\hline

\multirow{12}{*}{MNIST} 
  & \multirow{4}{*}{MLP} 
    & SAM & $\mathbf{0.9172 \pm 0.0269}$ \\
  & & GSAM & $\mathbf{0.9172 \pm 0.0201}$ \\
  & & GAM & $\mathbf{0.9172 \pm 0.0182}$ \\
  & & No regularizer & $0.9125 \pm 0.0212$ \\
\cline{2-4}
  & \multirow{4}{*}{MLP ReLU} 
    & SAM & $\mathbf{0.9469 \pm 0.0134}$ \\
  & & GSAM & $\mathbf{0.9313 \pm 0.0206}$ \\
  & & GAM & $\mathbf{0.9250 \pm 0.0175}$ \\
  & & No regularizer & $0.9187 \pm 0.0295$ \\
\cline{2-4}
  & \multirow{4}{*}{CNN} 
    & SAM & $0.9500 \pm 0.0351$ \\
  & & GSAM & $\mathbf{0.9797 \pm 0.0063}$ \\
  & & GAM & $\mathbf{0.9766 \pm 0.0148}$ \\
  & & No regularizer & $0.9734 \pm 0.0145$ \\
\hline

\multirow{12}{*}{Fashion-MNIST} 
  & \multirow{4}{*}{MLP} 
    & SAM & $\mathbf{0.8266 \pm 0.0329}$ \\
  & & GSAM & $\mathbf{0.8219 \pm 0.0294}$ \\
  & & GAM & $\mathbf{0.8156 \pm 0.0260}$ \\
  & & No regularizer & $0.8141 \pm 0.0310$ \\
\cline{2-4}
  & \multirow{4}{*}{MLP ReLU} 
    & SAM & $\mathbf{0.8078 \pm 0.0322}$ \\
  & & GSAM & $0.7797 \pm 0.0159$ \\
  & & GAM & $\mathbf{0.8172 \pm 0.0127}$ \\
  & & No regularizer & $0.7891 \pm 0.0644$ \\
\cline{2-4}
  & \multirow{4}{*}{CNN} 
    & SAM & $\mathbf{0.8438 \pm 0.0360}$ \\
  & & GSAM & $\mathbf{0.8734 \pm 0.0302}$ \\
  & & GAM & $\mathbf{0.8766 \pm 0.0302}$ \\
  & & No regularizer & $0.8141 \pm 0.1021$ \\
\hline

\multirow{12}{*}{CIFAR-10} 
  & \multirow{4}{*}{MLP} 
    & SAM & $0.3187 \pm 0.0234$ \\
  & & GSAM & $\mathbf{0.3781 \pm 0.0536}$ \\
  & & GAM & $\mathbf{0.3563 \pm 0.0269}$ \\
  & & No regularizer & $0.3391 \pm 0.0510$ \\
\cline{2-4}
  & \multirow{4}{*}{MLP ReLU} 
    & SAM & $0.3906 \pm 0.0422$ \\
  & & GSAM & $0.3172 \pm 0.0269$ \\
  & & GAM & $\mathbf{0.3984 \pm 0.0366}$ \\
  & & No regularizer & $0.3953 \pm 0.0460$ \\
\cline{2-4}
  & \multirow{4}{*}{CNN} 
    & SAM & $\mathbf{0.6062 \pm 0.0726}$ \\
  & & GSAM & $\mathbf{0.6188 \pm 0.0364}$ \\
  & & GAM & $\mathbf{0.6297 \pm 0.0460}$ \\
  & & No regularizer & $0.5750 \pm 0.0618$ \\
\hline

\end{tabular}%
}
\caption{Comparison of convergence accuracy across datasets, architectures, and regularizers using Vanilla LO, with and without learning regularization techniques.}
\label{tab:accuracy_results}
\end{table}

\begin{figure*}
    \centering
    \begin{tabular}{ccccc}
        \includegraphics[width=0.18\textwidth]{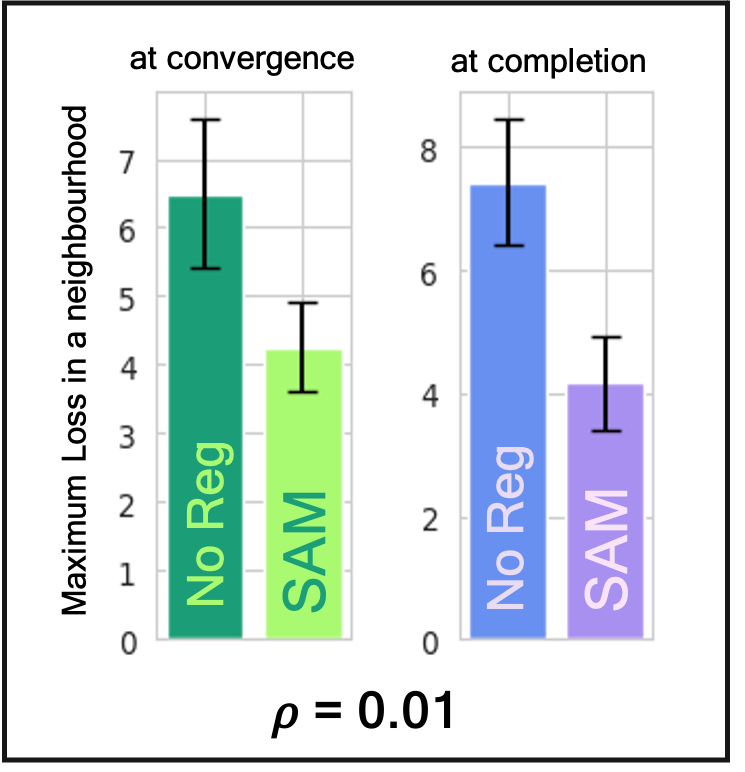} &
        \includegraphics[width=0.1785\textwidth]{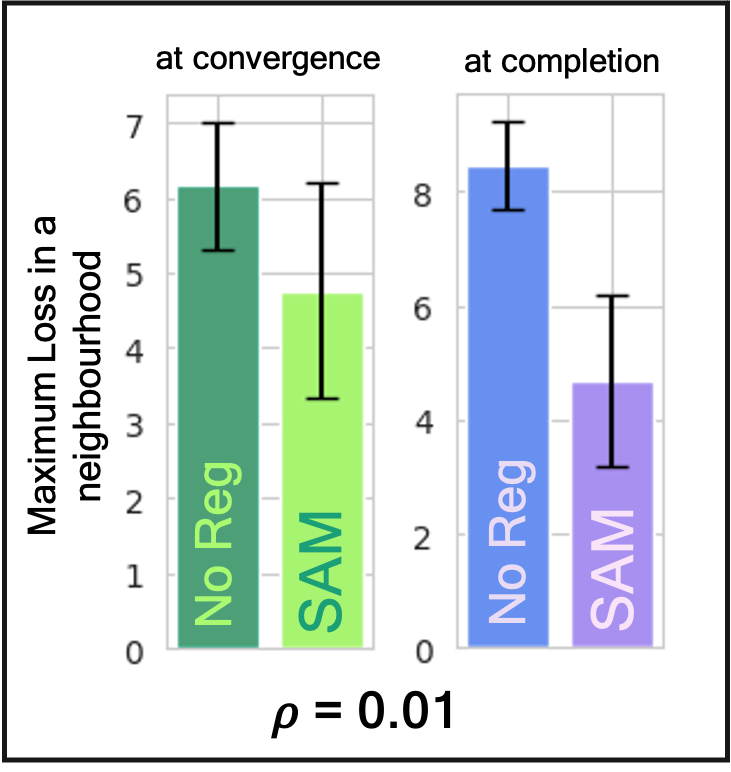} &
        \includegraphics[width=0.18\textwidth]{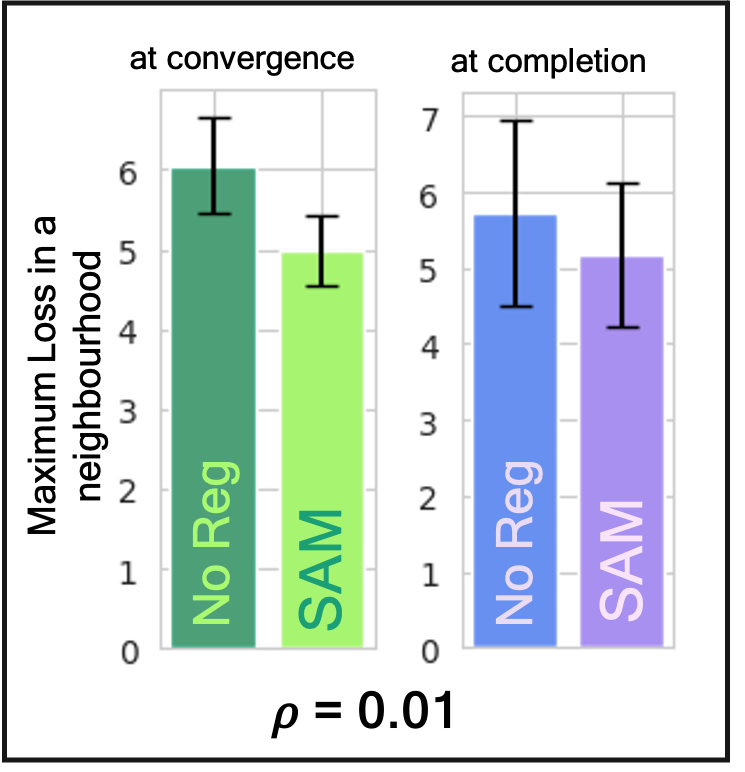} &
        \includegraphics[width=0.1875\textwidth]{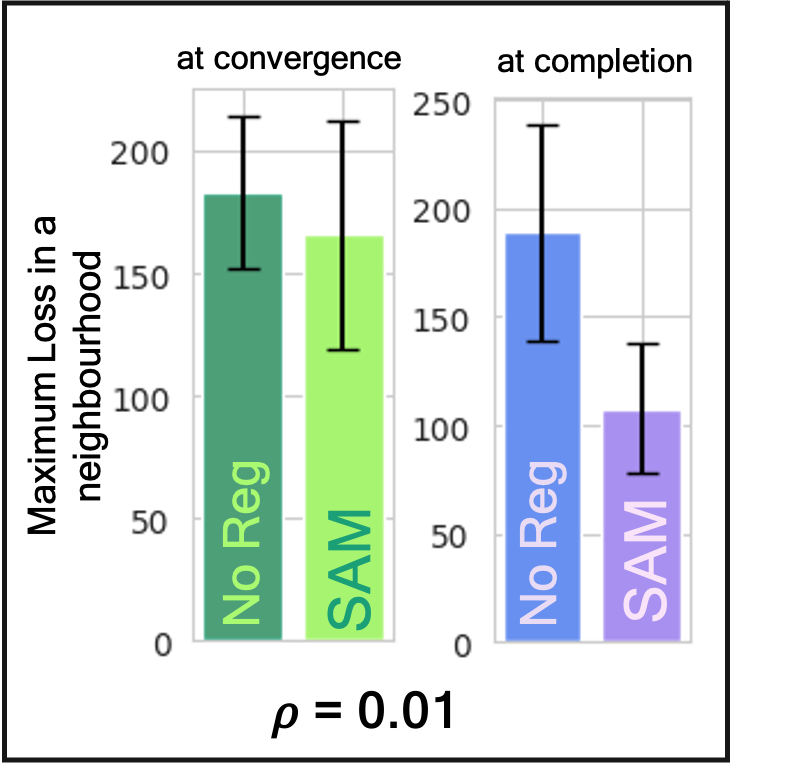} &
        \includegraphics[width=0.1875\textwidth]{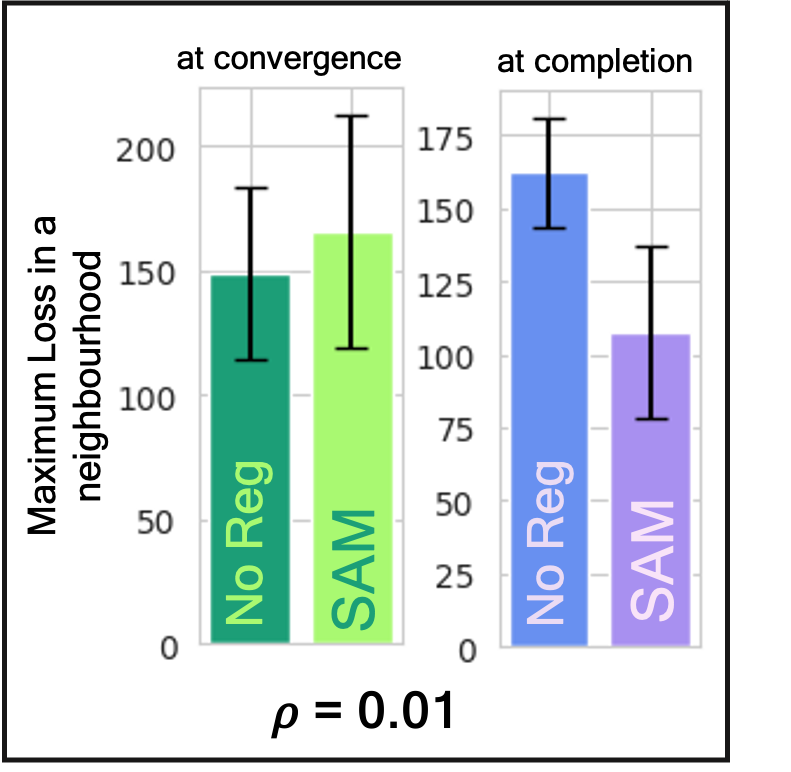} \\
        \scriptsize (a) MLP - MNIST & \scriptsize (b) MLP - FMNIST & \scriptsize (c) MLP - CIFAR-10 & 
        \scriptsize (d) MLP(Relu) - MNIST & \scriptsize (e) MLP(Relu) - FMNIST \\

    \end{tabular}
    \vspace{0.5em} 
    \begin{tabular}{cccc}
        \includegraphics[width=0.18\textwidth]{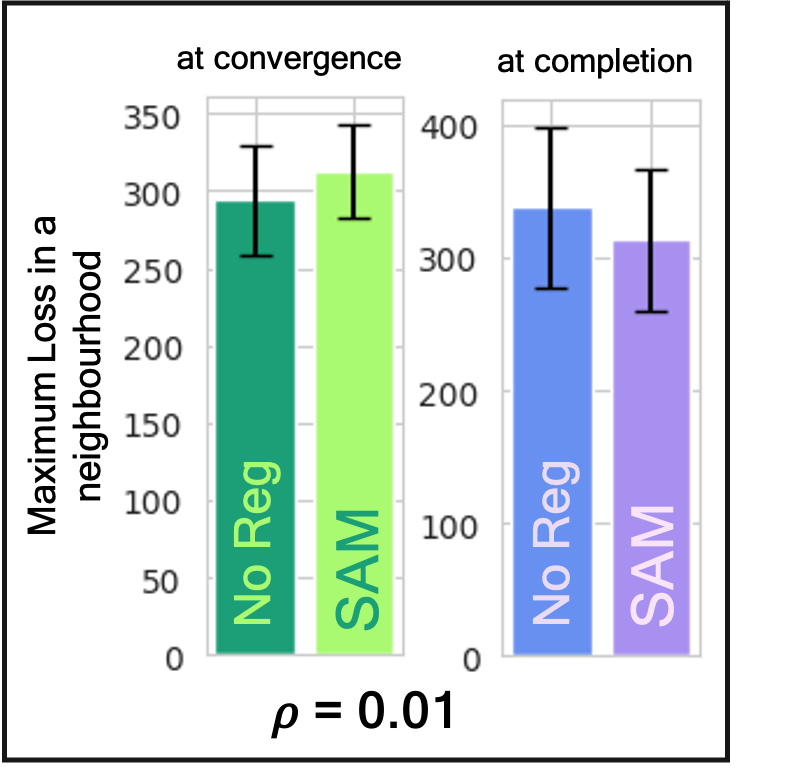} &
        \includegraphics[width=0.18\textwidth]{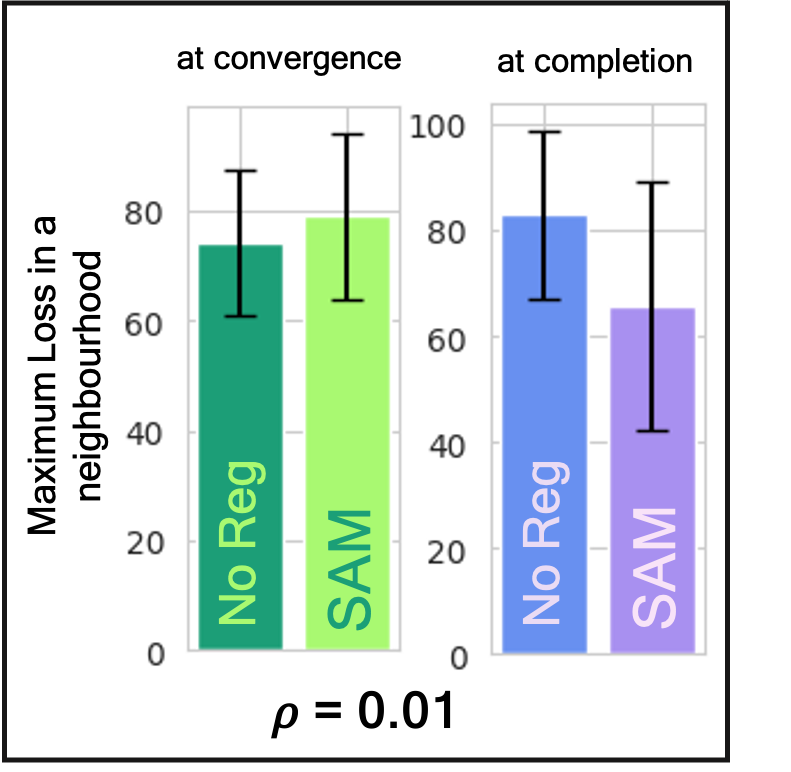} &
        \includegraphics[width=0.18\textwidth]{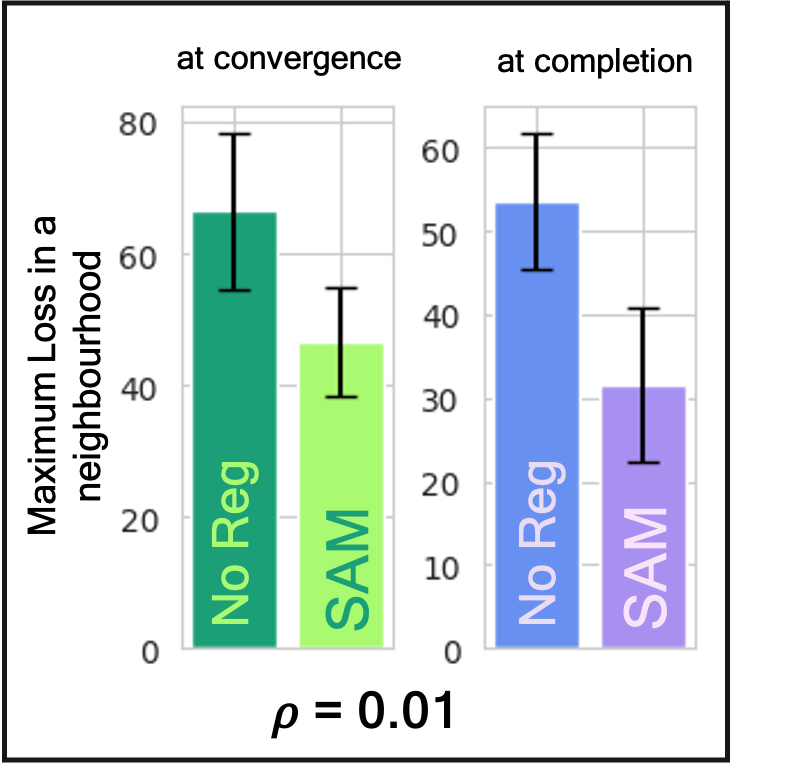} &
        \includegraphics[width=0.18\textwidth]{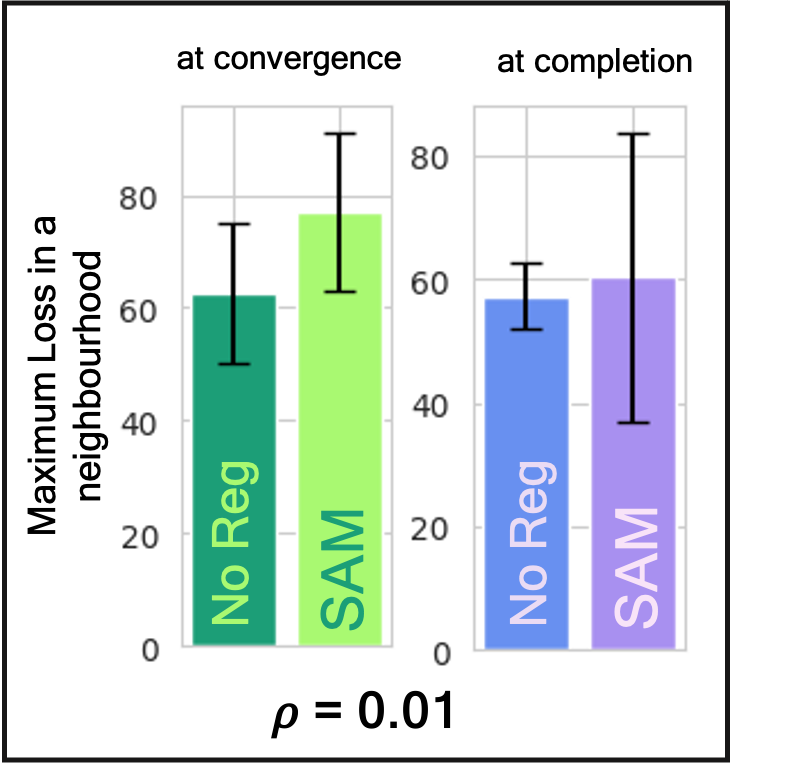} \\
        \scriptsize (f) MLP (ReLU) - CIFAR-10 & \scriptsize (g) CNN - MNIST & \scriptsize (h) CNN - FMNIST & \scriptsize (i) CNN - CIFAR-10 \\
    \end{tabular}
    
    \caption{Comparison of Vanilla LO with SAM-regularized LO across different architectures (MLP, MLP with ReLU, and CNN) and datasets (MNIST, FMNIST, CIFAR-10).}
    \label{fig:sam_all_tasks}
\end{figure*}

\begin{figure*}
    \centering
    \begin{tabular}{ccccc}
        \includegraphics[width=0.18\textwidth]{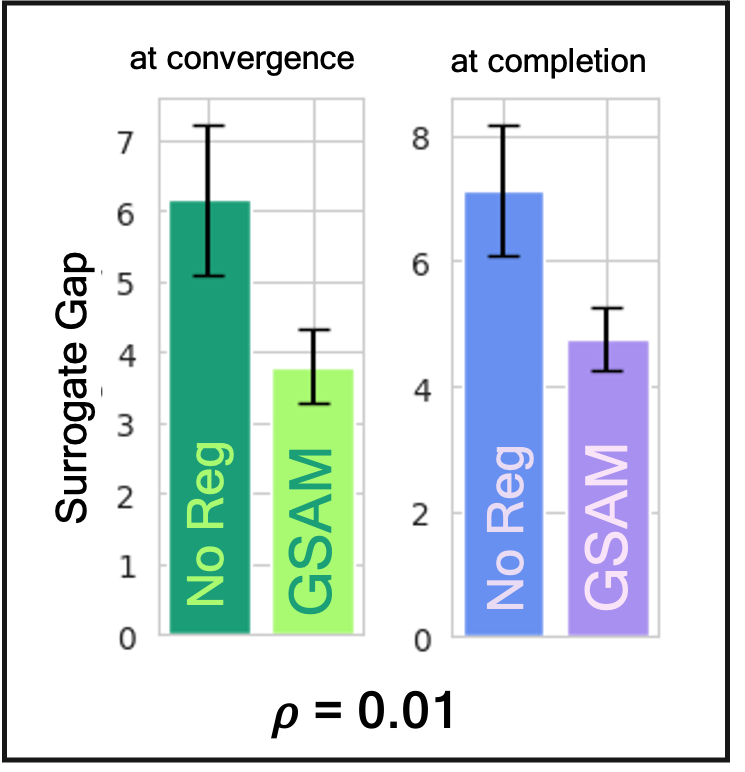} &
        \includegraphics[width=0.1785\textwidth]{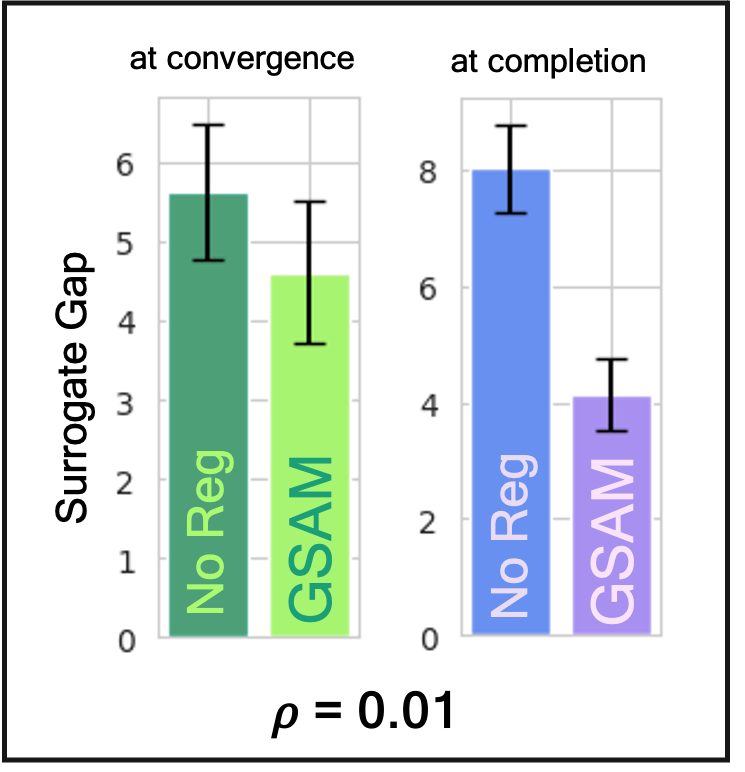} &
        \includegraphics[width=0.18\textwidth]{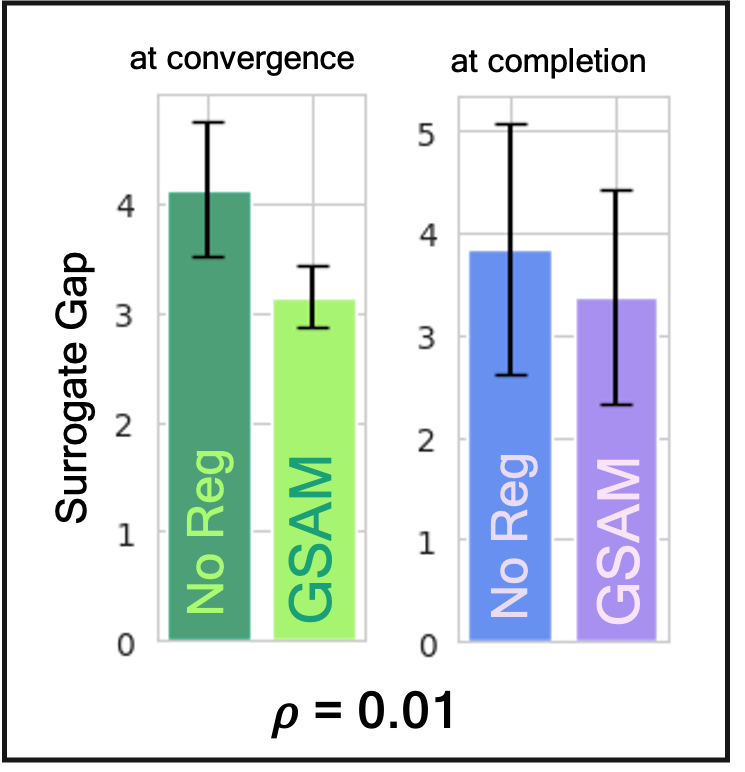} &
        \includegraphics[width=0.18\textwidth]{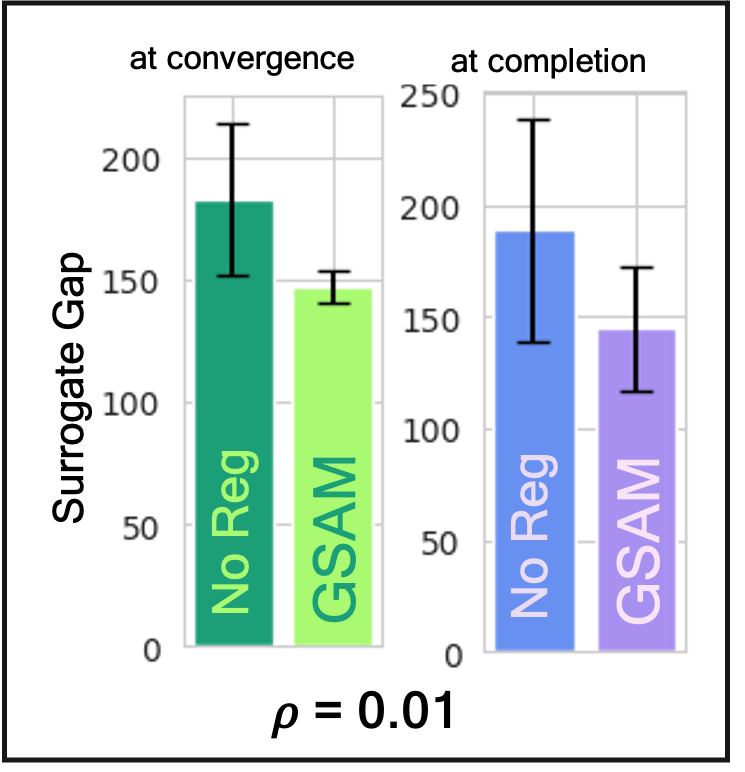} &
        \includegraphics[width=0.18\textwidth]{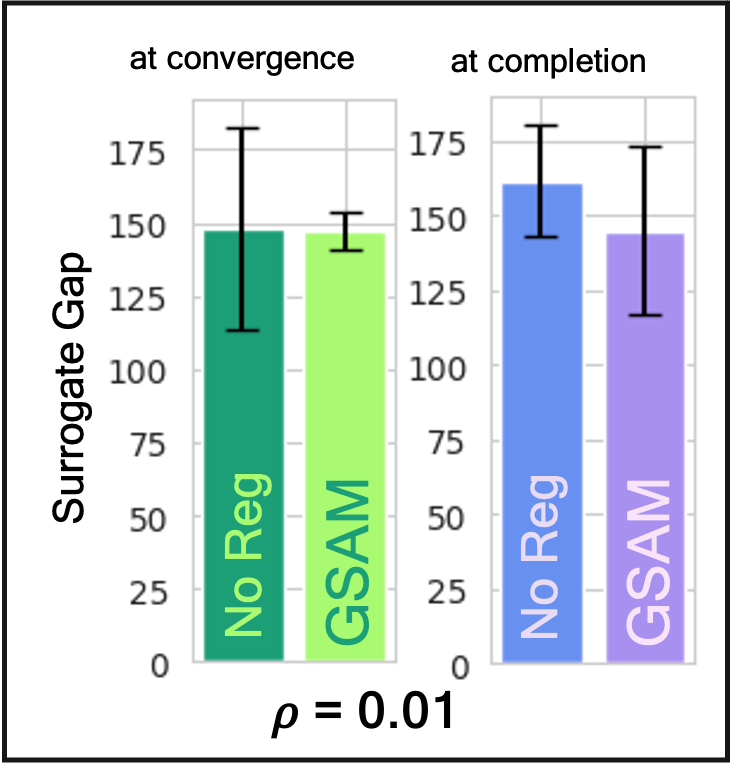} \\
        \scriptsize (a) MLP - MNIST & \scriptsize (b) MLP - FMNIST & \scriptsize (c) MLP - CIFAR-10 & 
        \scriptsize (d) MLP(ReLU) - MNIST & \scriptsize (e) MLP(ReLU) - FMNIST \\
    \end{tabular}
    
    \vspace{0.5em} 
    
    \begin{tabular}{cccc}
        \includegraphics[width=0.18\textwidth]{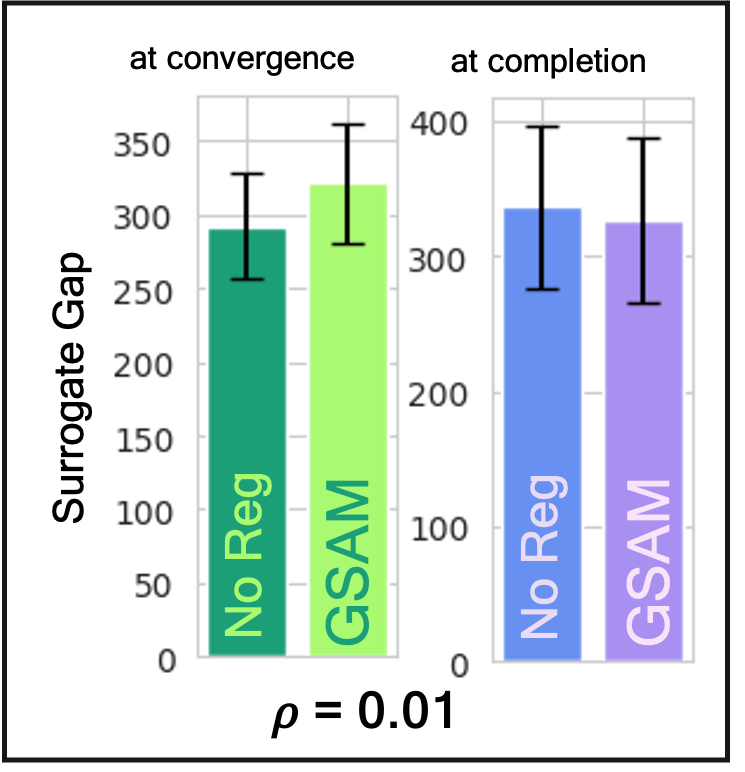} &
        \includegraphics[width=0.18\textwidth]{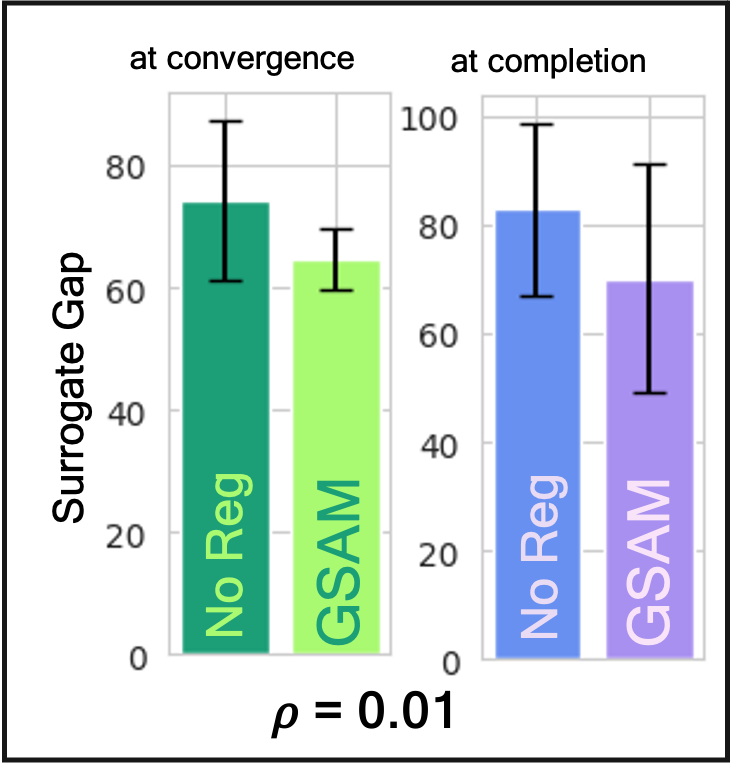} &
        \includegraphics[width=0.18\textwidth]{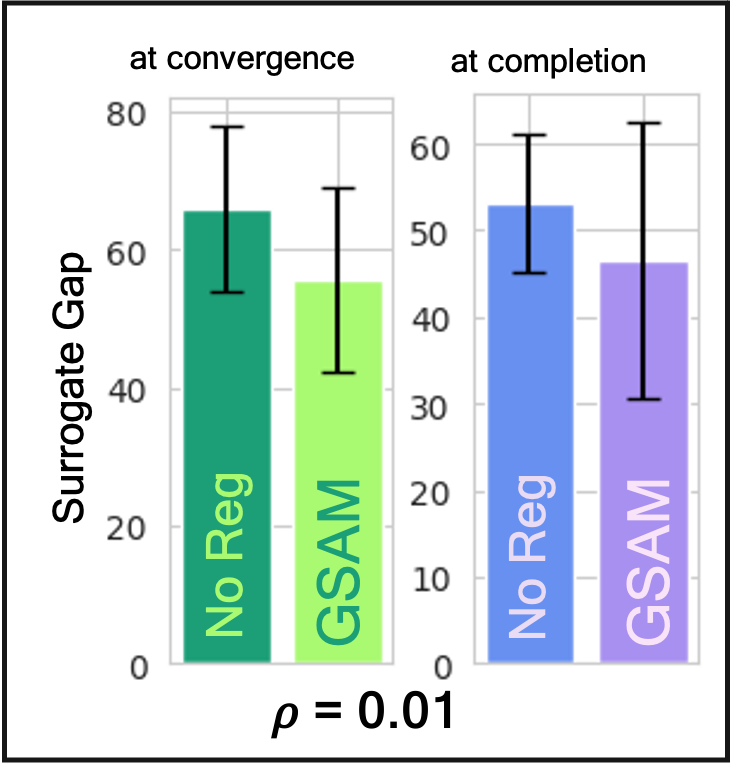} &
        \includegraphics[width=0.18\textwidth]{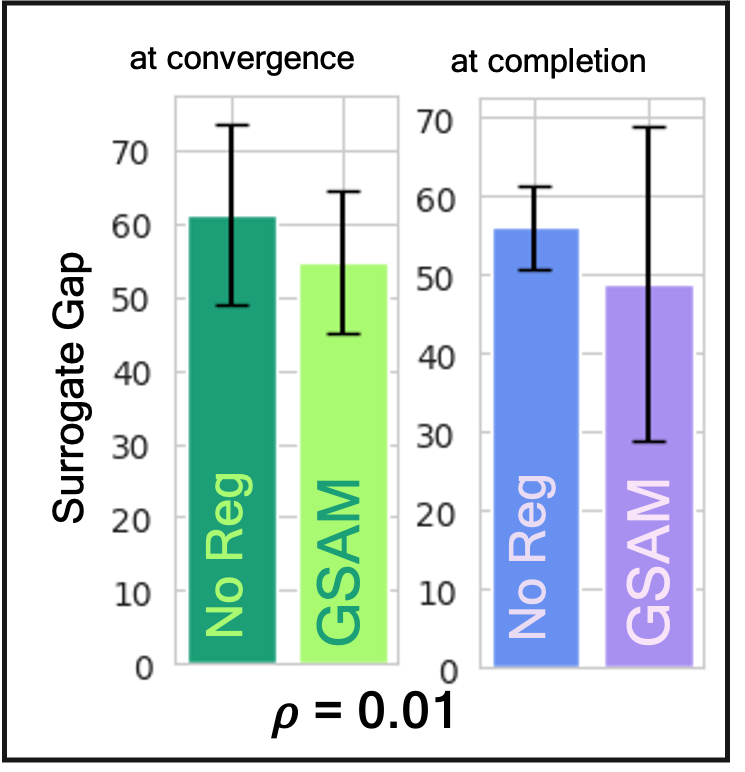} \\
        \scriptsize (f) MLP (ReLU) - CIFAR-10 & \scriptsize (g) CNN - MNIST & \scriptsize (h) CNN - FMNIST & \scriptsize (i) CNN - CIFAR-10 \\
    \end{tabular}
    
    \caption{Comparison of Vanilla LO with GSAM-regularized LO across different architectures (MLP, MLP with ReLU, and CNN) and datasets (MNIST, FMNIST, CIFAR-10).}
    \label{fig:gsam_all_tasks}
\end{figure*}

\begin{figure*}[htb]
    \centering
    \begin{tabular}{ccccc}
        \includegraphics[width=0.18\textwidth]{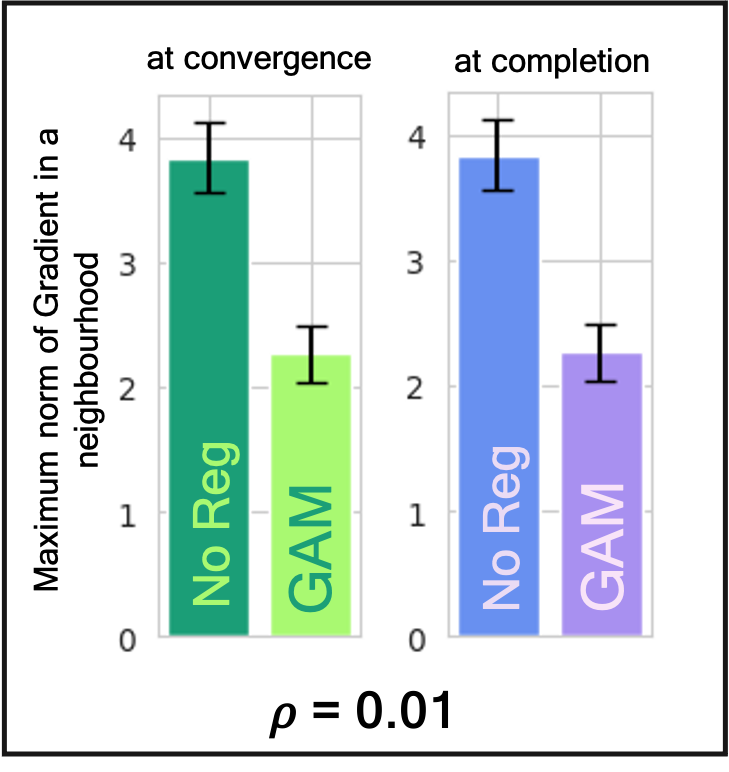} &
        \includegraphics[width=0.1785\textwidth]{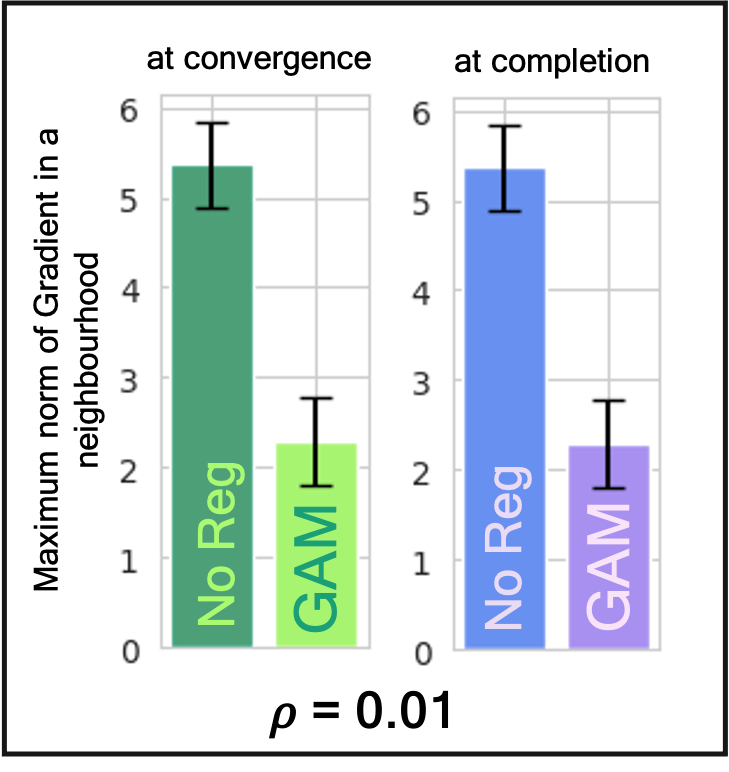} &
        \includegraphics[width=0.18\textwidth]{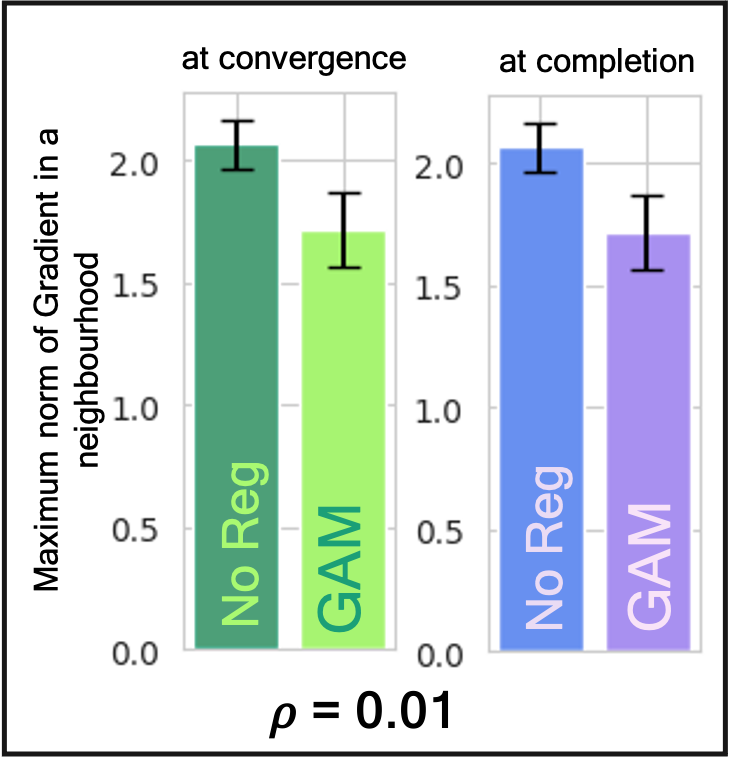} &
        \includegraphics[width=0.18\textwidth]{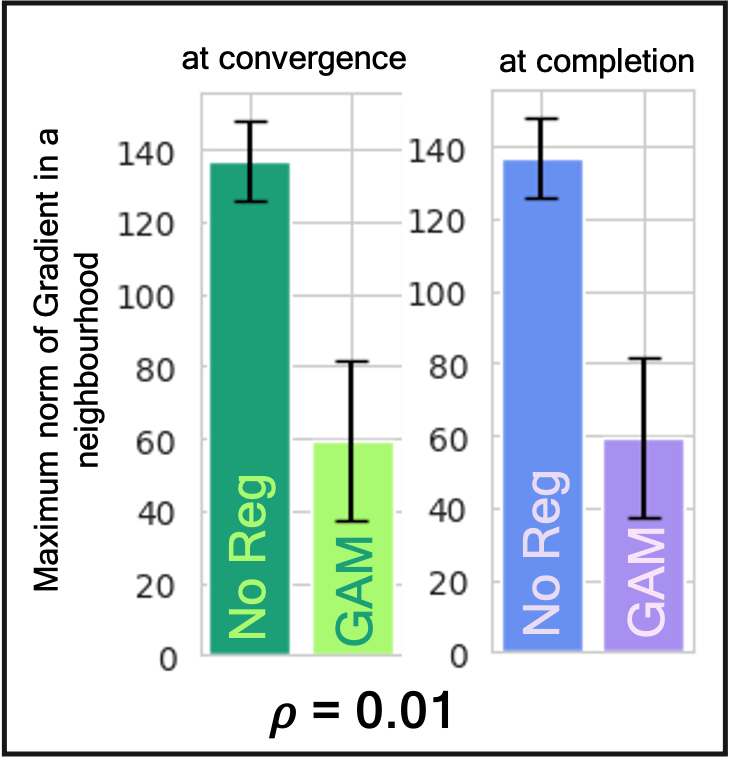} &
        \includegraphics[width=0.18\textwidth]{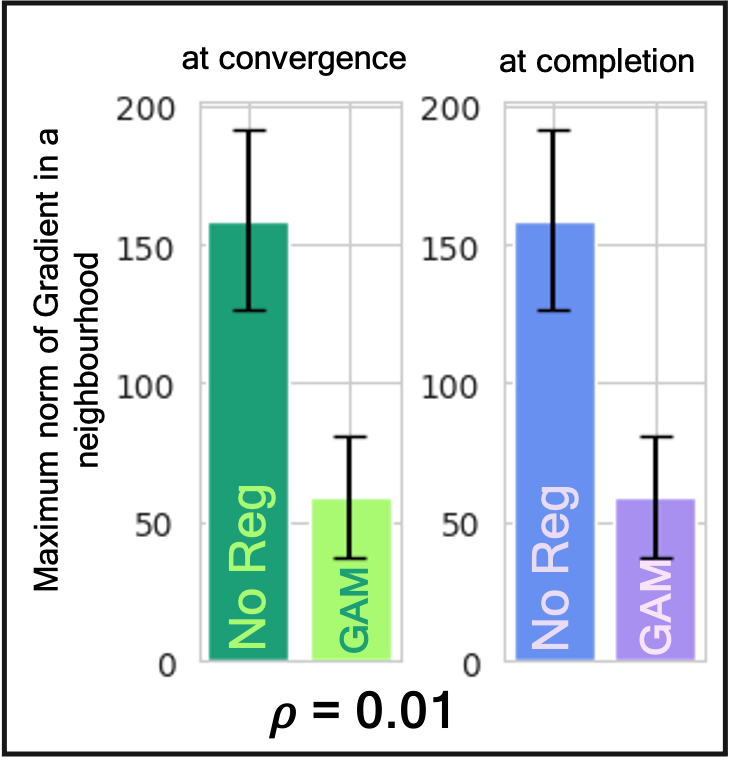} \\
        \scriptsize (a) MLP - MNIST & \scriptsize (b) MLP - FMNIST & \scriptsize (c) MLP - CIFAR-10 & 
        \scriptsize (d) MLP(ReLU) - MNIST & \scriptsize (e) MLP(ReLU) - FMNIST \\
    \end{tabular}
    
    \vspace{0.5em} 
    
    \begin{tabular}{cccc}
        \includegraphics[width=0.18\textwidth]{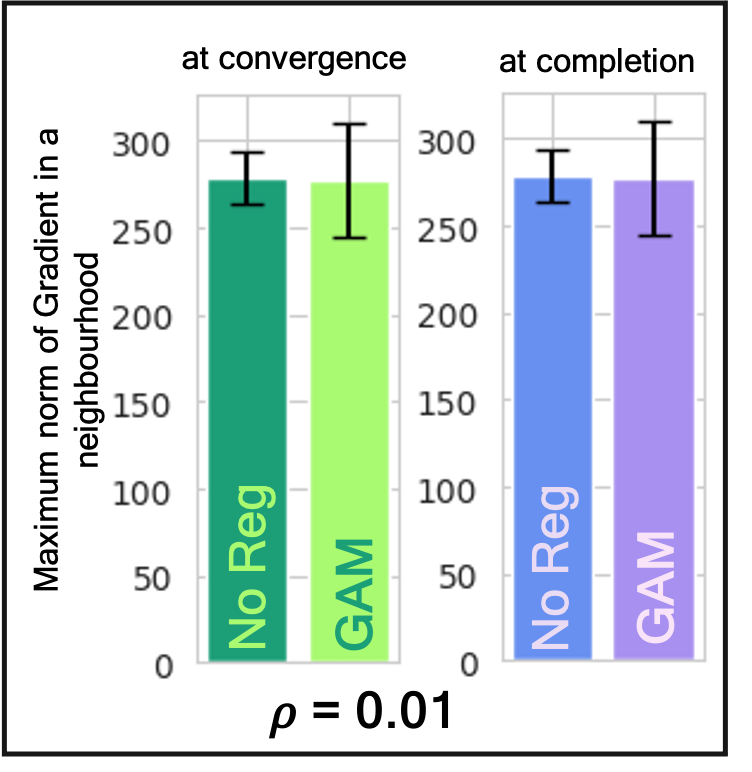} &
        \includegraphics[width=0.18\textwidth]{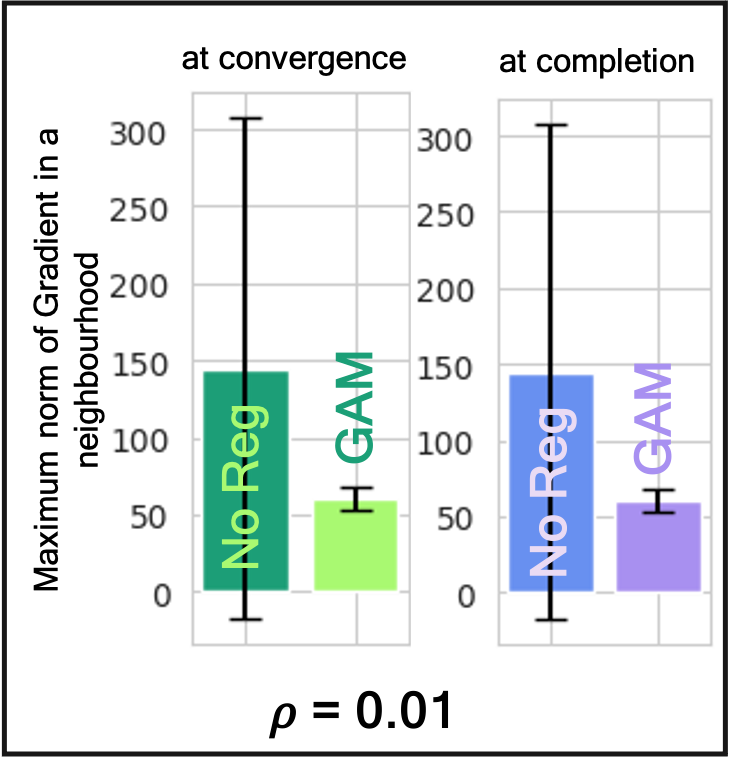} &
        \includegraphics[width=0.18\textwidth]{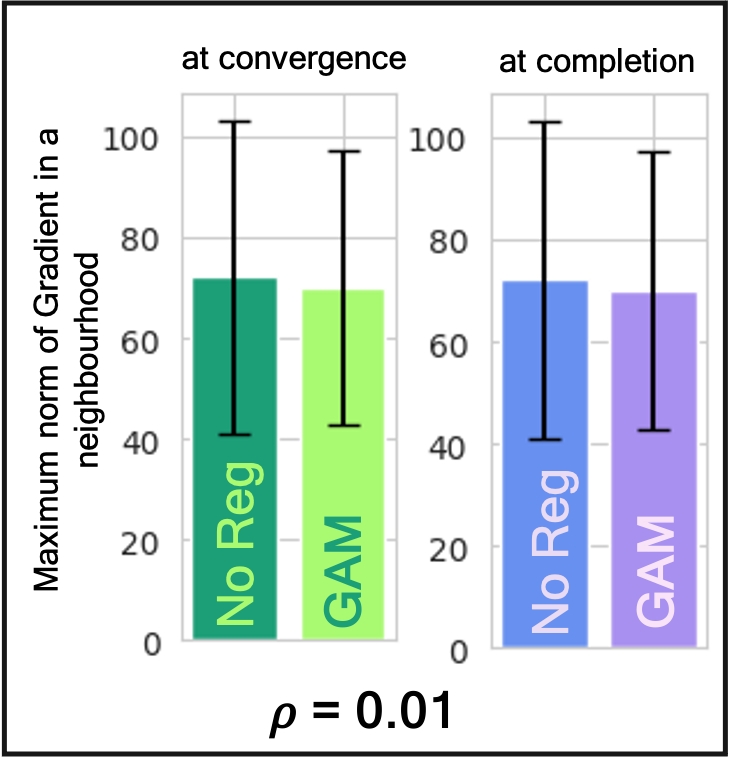} &
        \includegraphics[width=0.18\textwidth]{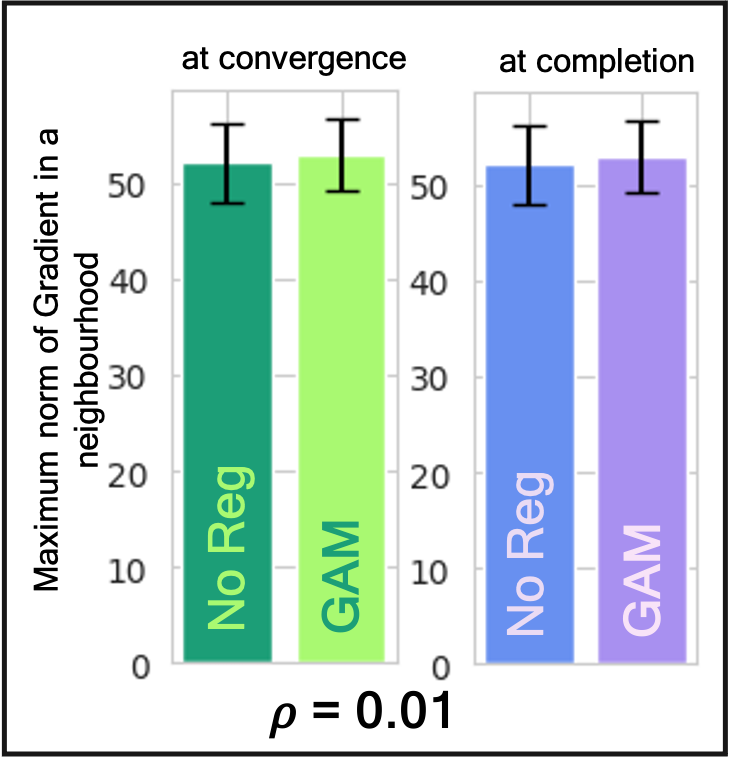} \\
        \scriptsize (f) MLP (ReLU) - CIFAR-10 & \scriptsize (g) CNN - MNIST & \scriptsize (h) CNN - FMNIST & \scriptsize (i) CNN - CIFAR-10 \\
    \end{tabular}
    
    \caption{Comparison of Vanilla LO with GAM-regularized LO across different architectures (MLP, MLP with ReLU, and CNN) and datasets (MNIST, FMNIST, CIFAR-10).}
    \label{fig:gam_all_tasks}
\end{figure*}

\subsection{Observations}
Our empirical results reveal distinct patterns in the behavior of the learned optimizer under different regularization techniques. In this section, we analyze key trends that emerged during evaluation, focusing on the learned optimizer's ability to reach points of convergence that are favored by the applied regularization technique. Specifically, we examine its ability to generalize across different landscapes by internalizing regularization strategies during training on a single landscape. 


Figures \ref{fig:sam_all_tasks}, \ref{fig:gsam_all_tasks}, \ref{fig:gam_all_tasks} present a comparative analysis of the different configurations under the hypothesis that a LO can learn the inherent property of SAM, GSAM and GAM respectively.


\begin{itemize}  
    \item \textbf{SAM:} SAM optimizes parameters within neighborhoods exhibiting uniformly low loss, thereby enhancing model generalizability and performance across diverse datasets and architectures. In MLP-based experiments (Figure~\ref{fig:sam_all_tasks} (a)-(c)), SAM effectively enforces the desired property, even when applied to CIFAR-10 (Figure~\ref{fig:sam_all_tasks} (c)), despite the significant deviation from the original task distribution. While this property may not always be distinctly observable at convergence due to the fixed patience iteration of 100, extended training yields PoCs with improved characteristics. However, for tasks that substantially differ from the original training objective, such as CIFAR-10 classification using a CNN, the validity of our hypothesis remains inconclusive.

    \item \textbf{GSAM:} Figure~\ref{fig:gsam_all_tasks} supports the hypothesis, demonstrating its validity either at convergence or upon completion. In MLP tasks with ReLU activation (Figure~\ref{fig:gsam_all_tasks} (d)-(f)), GSAM does not consistently follow the same trend as SAM and GAM but ultimately enforces the desired properties at completion. This discrepancy can be attributed to the distinct characteristics of PoC neighborhoods in MLP-ReLU models, as also observed in the loss profile of meta-training reported in \cite{aistat2023}. The challenges become more pronounced when applying GSAM to MLP(ReLU) + CIFAR-10, where increased task complexity and distribution shifts significantly influence the outcomes.

    \item \textbf{GAM:} The observed trends extend to CNN-based tasks, with CNN+MNIST (Figure~\ref{fig:gam_all_tasks} (g)) and CNN+FMNIST (Figure~\ref{fig:gam_all_tasks} (h)) demonstrating successful regularization. However, in CNN tasks involving CIFAR-10 (Figure~\ref{fig:gam_all_tasks} (i)), GAM fails to produce satisfactory results, likely due to the substantial distribution shift, which hinders the regularized LO from identifying optimal PoCs. Furthermore, across various experiments, the effectiveness of regularization is highly dependent on the choice of neighborhood radius. Extremely small radii may fail to capture meaningful variations, whereas excessively large radii risk encompassing multiple minima, making property evaluation unreliable. A neighborhood radius of 0.01 is suggested as a balanced choice to mention here, though its suitability may vary depending on the loss landscape.
\end{itemize}  


\section{Conclusion}
Incorporating these advanced regularization techniques into the training of neural optimizers offers a promising pathway to enhance their performance. By embedding the principles of SAM, GSAM, and GAM into the optimizer's learning process, the neural optimizer can develop an intrinsic understanding of the loss landscape's geometry. This internalization enables the optimizer to navigate towards flatter and more generalizable minima without the need for explicit regularization during the training of the optimizee.  

Furthermore, the LSTM in the learned optimizer (LO) attempts to understand the trajectory leading to the PoC and generalizes this understanding across different loss surfaces. By capturing the structural patterns of one loss surface, it can leverage this knowledge to optimize even dissimilar surfaces, provided they exhibit analogous structural properties. This capability suggests that training on multiple types of loss surfaces can enhance the optimizer's ability to generalize, potentially improving convergence across diverse tasks.  

This strategy not only streamlines the optimization process by reducing computational overhead associated with external regularization but also fosters the development of models with improved generalization capabilities. By learning to anticipate and adjust for potential sharpness in the loss landscape, the neural optimizer becomes adept at guiding the training process towards solutions that are both robust and efficient.  

However, hyperparameter tuning remains an essential factor in further refining performance. Optimally selecting training configurations, especially when incorporating multiple types of loss surfaces, could significantly enhance the optimizer’s effectiveness in navigating complex landscapes.  

In summary, the integration of these regularization methodologies into neural optimizers, combined with structured training across diverse loss surfaces, represents a significant advancement in machine learning. This approach offers a cohesive framework for achieving enhanced generalization and performance in trained models.  

\section*{Acknowledgements}
 CK and SKS gratefully acknowledge Hrithik Suresh for the discussions related to the work and CK and SKS thank IIT Palakkad for the access to Param Vidhya and CDAC for Param Siddhi AI Cluster.

\bibliography{sample}

\begin{thebibliography}{21}
\providecommand{\natexlab}[1]{#1}

\bibitem[{Almeida et~al.(2021)Almeida, Winter, Tang, and Zaremba}]{optimizerGen}
Almeida, D.; Winter, C.; Tang, J.; and Zaremba, W. 2021.
\newblock A generalizable approach to learning optimizers.
\newblock \emph{arXiv preprint arXiv:2106.00958}.

\bibitem[{Andrychowicz et~al.(2016)Andrychowicz, Denil, Gomez, Hoffman, Pfau, Schaul, Shillingford, and De~Freitas}]{L2LGDGD}
Andrychowicz, M.; Denil, M.; Gomez, S.; Hoffman, M.~W.; Pfau, D.; Schaul, T.; Shillingford, B.; and De~Freitas, N. 2016.
\newblock Learning to learn by gradient descent by gradient descent.
\newblock \emph{Advances in neural information processing systems}, 29.

\bibitem[{Ba, Kiros, and Hinton(2016)}]{layer_normalization}
Ba, J.~L.; Kiros, J.~R.; and Hinton, G.~E. 2016.
\newblock Layer normalization.
\newblock \emph{arXiv preprint arXiv:1607.06450}.

\bibitem[{Chen et~al.(2020)Chen, Zhang, Jingyang, Chang, Liu, Amini, and Wang}]{Curriculum_chen2020}
Chen, T.; Zhang, W.; Jingyang, Z.; Chang, S.; Liu, S.; Amini, L.; and Wang, Z. 2020.
\newblock Training stronger baselines for learning to optimize.
\newblock \emph{Advances in Neural Information Processing Systems}, 33: 7332--7343.

\bibitem[{Foret et~al.(2020)Foret, Kleiner, Mobahi, and Neyshabur}]{sam}
Foret, P.; Kleiner, A.; Mobahi, H.; and Neyshabur, B. 2020.
\newblock Sharpness-aware minimization for efficiently improving generalization.
\newblock \emph{arXiv preprint arXiv:2010.01412}.

\bibitem[{Franceschi et~al.(2018)Franceschi, Frasconi, Salzo, Grazzi, and Pontil}]{bilevel}
Franceschi, L.; Frasconi, P.; Salzo, S.; Grazzi, R.; and Pontil, M. 2018.
\newblock Bilevel programming for hyperparameter optimization and meta-learning.
\newblock In \emph{International conference on machine learning}, 1568--1577. PMLR.

\bibitem[{He, Huang, and Yuan(2019)}]{asym_valley}
He, H.; Huang, G.; and Yuan, Y. 2019.
\newblock Asymmetric valleys: Beyond sharp and flat local minima.
\newblock \emph{Advances in neural information processing systems}, 32.

\bibitem[{Hoerl and Kennard(1970)}]{L2_reg_ridge}
Hoerl, A.~E.; and Kennard, R.~W. 1970.
\newblock Ridge regression: applications to nonorthogonal problems.
\newblock \emph{Technometrics}, 12(1): 69--82.

\bibitem[{Ioffe and Szegedy(2015)}]{batch_norm}
Ioffe, S.; and Szegedy, C. 2015.
\newblock Batch normalization: Accelerating deep network training by reducing internal covariate shift.
\newblock In \emph{International conference on machine learning}, 448--456. pmlr.

\bibitem[{Li et~al.(2018)Li, Xu, Taylor, Studer, and Goldstein}]{2018visualizinglosslandscape}
Li, H.; Xu, Z.; Taylor, G.; Studer, C.; and Goldstein, T. 2018.
\newblock Visualizing the loss landscape of neural nets.
\newblock \emph{Advances in neural information processing systems}, 31.

\bibitem[{Maheswaranathan et~al.(2021)Maheswaranathan, Sussillo, Metz, Sun, and Sohl-Dickstein}]{maheswaranathan2021reverse}
Maheswaranathan, N.; Sussillo, D.; Metz, L.; Sun, R.; and Sohl-Dickstein, J. 2021.
\newblock Reverse engineering learned optimizers reveals known and novel mechanisms.
\newblock \emph{Advances in neural information processing systems}, 34: 19910--19922.

\bibitem[{Metz et~al.(2020)Metz, Maheswaranathan, Freeman, Poole, and Sohl-Dickstein}]{hierarchyMLP}
Metz, L.; Maheswaranathan, N.; Freeman, C.~D.; Poole, B.; and Sohl-Dickstein, J. 2020.
\newblock Tasks, stability, architecture, and compute: Training more effective learned optimizers, and using them to train themselves.
\newblock \emph{arXiv preprint arXiv:2009.11243}.

\bibitem[{Metz et~al.(2019)Metz, Maheswaranathan, Nixon, Freeman, and Sohl-Dickstein}]{mlpLO}
Metz, L.; Maheswaranathan, N.; Nixon, J.; Freeman, D.; and Sohl-Dickstein, J. 2019.
\newblock Understanding and correcting pathologies in the training of learned optimizers.
\newblock In \emph{International Conference on Machine Learning}, 4556--4565. PMLR.

\bibitem[{Perez and Wang(2017)}]{data_augmentation}
Perez, L.; and Wang, J. 2017.
\newblock The effectiveness of data augmentation in image classification using deep learning.
\newblock \emph{arXiv preprint arXiv:1712.04621}.

\bibitem[{Srivastava et~al.(2014)Srivastava, Hinton, Krizhevsky, Sutskever, and Salakhutdinov}]{dropout}
Srivastava, N.; Hinton, G.; Krizhevsky, A.; Sutskever, I.; and Salakhutdinov, R. 2014.
\newblock Dropout: a simple way to prevent neural networks from overfitting.
\newblock \emph{The journal of machine learning research}, 15(1): 1929--1958.

\bibitem[{Tibshirani(1996)}]{l1_reg_tibshirani}
Tibshirani, R. 1996.
\newblock Regression shrinkage and selection via the lasso.
\newblock \emph{Journal of the Royal Statistical Society Series B: Statistical Methodology}, 58(1): 267--288.

\bibitem[{Wichrowska et~al.(2017)Wichrowska, Maheswaranathan, Hoffman, Colmenarejo, Denil, Freitas, and Sohl-Dickstein}]{hierarchyRNN}
Wichrowska, O.; Maheswaranathan, N.; Hoffman, M.~W.; Colmenarejo, S.~G.; Denil, M.; Freitas, N.; and Sohl-Dickstein, J. 2017.
\newblock Learned optimizers that scale and generalize.
\newblock In \emph{International conference on machine learning}, 3751--3760. PMLR.

\bibitem[{Xiong and Hsieh(2022)}]{smoothDM}
Xiong, Y.; and Hsieh, C.-J. 2022.
\newblock Learning to Learn with Smooth Regularization.
\newblock In Avidan, S.; Brostow, G.; Ciss{\'e}, M.; Farinella, G.~M.; and Hassner, T., eds., \emph{Computer Vision -- ECCV 2022}, 550--565. Cham: Springer Nature Switzerland.

\bibitem[{Yang et~al.(2023)Yang, Chen, Zhu, He, Tao, Liang, and Wang}]{aistat2023}
Yang, J.; Chen, T.; Zhu, M.; He, F.; Tao, D.; Liang, Y.; and Wang, Z. 2023.
\newblock Learning to Generalize Provably in Learning to Optimize.
\newblock In \emph{International Conference on Artificial Intelligence and Statistics}, 9807--9825. PMLR.

\bibitem[{Zhang et~al.(2023)Zhang, Xu, Yu, Zou, and Cui}]{gam}
Zhang, X.; Xu, R.; Yu, H.; Zou, H.; and Cui, P. 2023.
\newblock Gradient norm aware minimization seeks first-order flatness and improves generalization.
\newblock In \emph{Proceedings of the IEEE/CVF Conference on Computer Vision and Pattern Recognition}, 20247--20257.

\bibitem[{Zhuang et~al.(2022)Zhuang, Gong, Yuan, Cui, Adam, Dvornek, Tatikonda, Duncan, and Liu}]{gsam}
Zhuang, J.; Gong, B.; Yuan, L.; Cui, Y.; Adam, H.; Dvornek, N.; Tatikonda, S.; Duncan, J.; and Liu, T. 2022.
\newblock Surrogate gap minimization improves sharpness-aware training.
\newblock \emph{arXiv preprint arXiv:2203.08065}.

\end{thebibliography}

\newpage
\onecolumn

\noindent \hrulefill
\section*{Supplementary Material: \\ Learning Regularizers: Learning Optimizers that can Regularize}

\noindent \hrulefill
\vspace{1em}

This supplementary section outlines the experimental setup and provides insights into the process of learning a regularizer through a learned optimizer (LO). We describe how the LO is trained on optimizee tasks in the presence of regularization and how it captures the effect of different regularizers through gradient based updates. The section also complements previously presented results, particularly early evidence from $L_2$-regularized regression, and includes extended evaluations across various datasets and neural network architectures to support our hypothesis.

\section{Learning a Regularizer}\label{appendix:learn_reg}

The previous sections have introduced neural optimizers capable of capturing the desired properties of various regularization techniques. Achieving this capability necessitates a specialized training algorithm. We developed a strategy that adapts neural optimizer training by incorporating a penalty into the optimizer's loss function, effectively acting as a regularizer for its updates. This penalty is derived from the specific property that the regularization method aims to exploit.

\begin{figure}[!htb]
    \centering
    \includegraphics[width=0.7\textwidth]{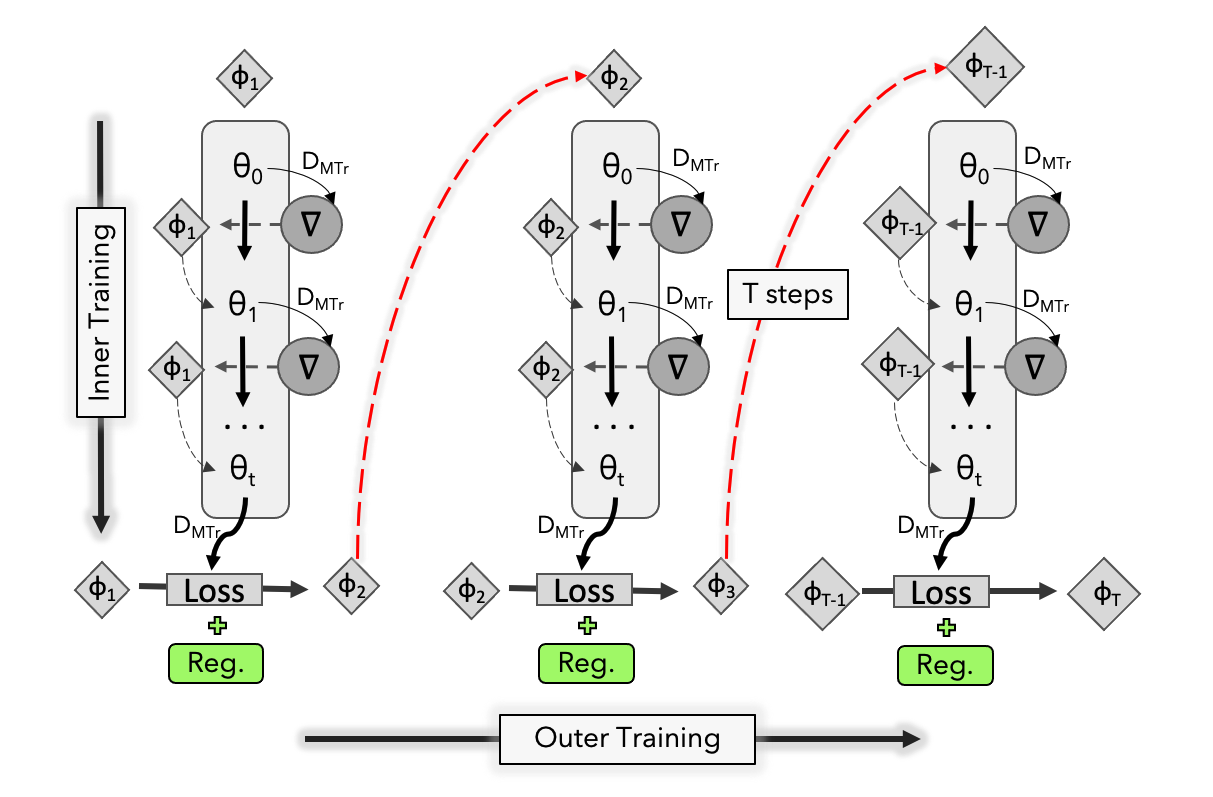}
    \caption{Learning a Regularizer}
    \label{fig:LO_training_reg}
\end{figure}

For instance, consider the use of the Hessian as a regularizer, as explored in \cite{aistat2023}. In this approach, the Hessian is computed at the point of convergence (PoC), and the optimizer is penalized based on a chosen norm of the Hessian. This guides the optimizer's updates toward PoCs characterized by a lower Hessian norm, which is presumably indicative of flatter minima. Figure \ref{fig:LO_training_reg} illustrates the step in which the regularization term is added, depicting the overall structure of the training process.

In our study, we examine newer and more promising regularization techniques to assess whether a learned optimizer (LO) can internalize the geometric properties these techniques target. While our strategy aligns with previous methods, penalizing the LO using these advanced regularization techniques involves more complexity than merely calculating the Hessian on the optimizee's loss surface, \( l(\theta_t(\phi)) \).

Take, for example, Sharpness-Aware Minimization (SAM) (detailed in Section \ref{sec:SAM}). In SAM, the gradient of the regularized loss with respect to the optimizer's parameters is approximated by the gradient at a perturbed parameter set, \( \phi^{adv} \). This gradient can only be obtained after performing Truncated Backpropagation Through Time (TBPTT) steps on the LO's parameters, starting from the same state of the optimizee. This forward pass over multiple TBPTT steps is crucial for guiding the LO's training process, thereby linking the optimizee's loss to the regularization applied to the optimizer's loss.

Consequently, when SAM is applied to the optimizer's loss, it computes gradients with respect to the optimizer's parameters (\( \phi \)) and updates these parameters to enhance the optimizer's effectiveness in minimizing the regularized loss.

\textbf{Note}: In all our experiments, the optimizer's loss is defined as the loss obtained from the optimizee's state at the final step of TBPTT, rather than the cumulative loss over all TBPTT steps. Our experiments indicated that this approach led to better convergence compared to the latter method. A similar strategy is adopted for other regularizers, including GSAM and GAM.

\section{Training an Optimizee}

This phase, also known as the \textbf{Meta-Testing Phase}, consists of two steps: Meta-Test Training and Meta-Test Testing. Meta-Test Training is analogous to a single step of inner training, where a pre-trained optimizer is used to train an optimizee task. The data for Meta-Testing is divided into two parts: Meta-Test Train Data (\( D_{MTsTr} \)) and Meta-Test Test Data (\( D_{MTsTs} \)). We selected a set of optimizee tasks and evaluated all our trained LOs on them for subsequent analysis.

\begin{figure}[!htb]
    \centering
    \includegraphics[width=0.5\textwidth]{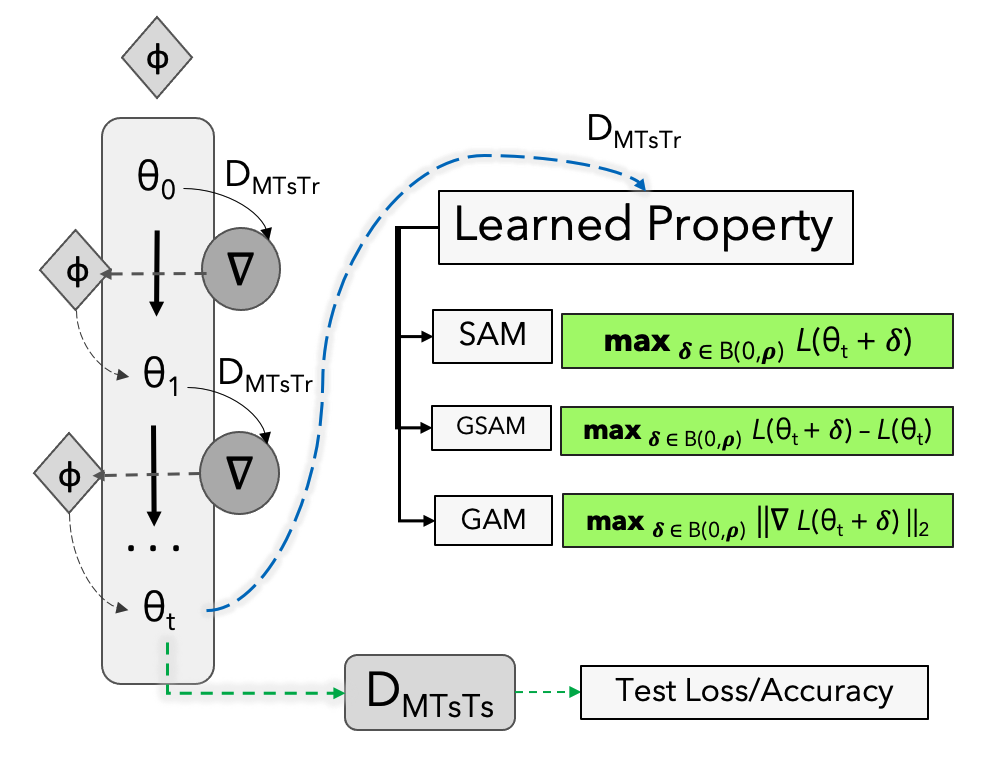}
    \caption{Training an Optimizee/Task using an LO}
    \label{fig:opzee_training}
\end{figure}

The tasks include:

\begin{itemize}
    \item MNIST/Fashion MNIST/CIFAR10 classification using a \textbf{Multilayer Perceptron (MLP)} with a single layer of 20 neurons and \textbf{sigmoid} activation function.
    \item MLP with a single layer of 20 neurons and \textbf{ReLU} activation function.
    \item \textbf{Convolutional Neural Network (CNN)} comprising two 2D convolutional layers with 16 \( 3\times3 \) kernels and 32 \( 5\times5 \) kernels, respectively, with a 2D max-pooling layer (kernel size \( 2\times2 \)) in between.
\end{itemize}

\section{Early Evidence: Learned Optimizers Can Regularize}

\subsection{Experimental Setup}

As an initial investigation into our hypothesis that a learned optimizer (LO) can inherently exhibit regularization behavior, we conducted a simple regression experiment with explicit $L_2$ regularization. This served as a proof-of-concept to probe whether the optimizer could internalize regularization during training.

To evaluate this, we designed a controlled setup based on a regression task with $L_2$ regularization. The experimental procedure consisted of the following steps:

\begin{itemize}
    \item \textbf{Meta-Training of the Learned Optimizer:} The LO was meta-trained on a distribution of randomly generated third-degree polynomials with added noise. This training aimed to equip the optimizer with the ability to generalize across similar regression tasks.
    
    \item \textbf{Meta-Test and Baseline Training:} A new set of third-degree polynomial datasets was used for evaluating the performance of the learned optimizer and comparing it against standard stochastic gradient descent (SGD).
    
    \item \textbf{Training Configuration:} Both the LO (during meta-testing) and the SGD baseline were trained using identical datasets and settings to ensure a fair comparison:
    \begin{itemize}
        \item \textbf{LO Meta-Test Training:} 5000 epochs with early stopping using a patience of 500 epochs.
        \item \textbf{SGD Training:} Standard gradient descent with the same number of epochs and early stopping criteria. For regularized runs, weight decay was set to 0.1.
        \item \textbf{Evaluation:} After training, both models were evaluated on a test set using the $R^2$ score. Additionally, the $L_2$-norm of the learned parameters was computed to assess the extent of regularization.
    \end{itemize}
\end{itemize}

\subsection{Results and Observations}
\begin{figure}[htbp]
    \centering
    \includegraphics[width=0.95\linewidth]{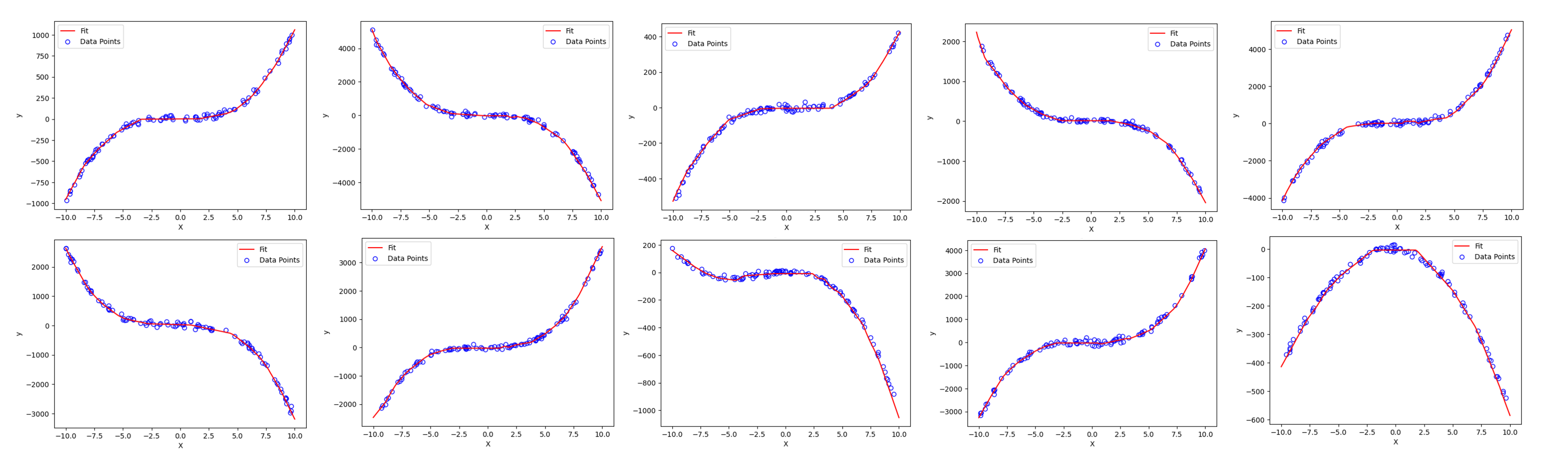}
    \caption{Meta Testing on regression task using LO trained without regularized Meta-training loss. 
    \textbf{R² scores:} [0.9975, 0.9939, 0.9932, 0.9978, 0.9937, 0.9962, 0.9969, 0.9883, 0.9961, 0.9892]. 
    \textbf{Norm values:} [1129.36, 6176.41, 894.11, 2362.60, 4252.39, 8152.70, 3160.35, 1696.55, 3822.35, 670.14].}
    \label{fig:DMO_no_regn_regression}
\end{figure}

\begin{figure}[htbp]
    \centering
    \includegraphics[width=0.95\linewidth]{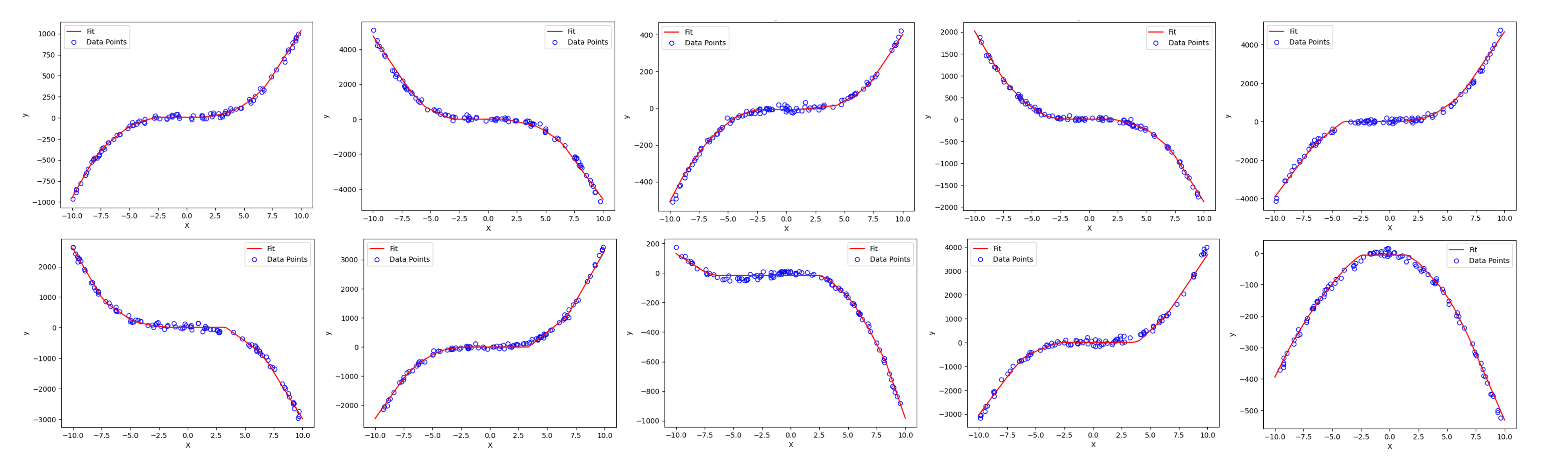}
    \caption{Meta Testing on regression task using LO trained with regularized Meta-training loss. 
    \textbf{R² scores:} [0.9956, 0.9540, 0.9953, 0.9840, 0.8511, 0.9694, 0.9756, 0.9946, 0.7574, 0.9944]. 
    \textbf{Norm values:} [255.51, 1921.37, 145.69, 527.79, 2015.03, 1116.32, 1411.17, 409.56, 2373.15, 120.06].}
    \label{fig:DMO_regn_regression}
\end{figure}

\begin{figure}[htbp]
    \centering
    \includegraphics[width=0.95\linewidth]{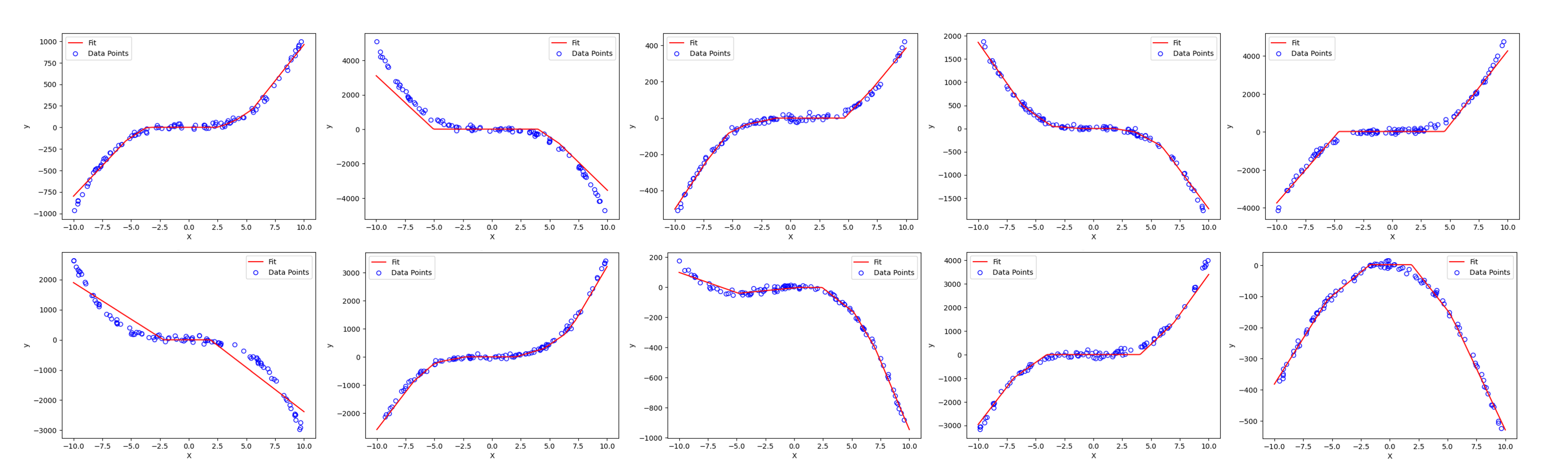}
    \caption{Testing on a regression task using the SGD optimizer. 
    \textbf{R² scores:} [0.9851, 0.9450, 0.9922, 0.9916, 0.9852, 0.5425, 0.9923, 0.9946, 0.9829, 0.9943]. 
    \textbf{Norm values:} [101.81, 2310.26, 54.68, 271.15, 1846.31, 1663.45, 958.72, 70.66, 1383.07, 32.82].}
    \label{fig:SGD_no_regn_regression}
\end{figure}

\begin{figure}[htbp]
    \centering
    \includegraphics[width=0.95\linewidth]{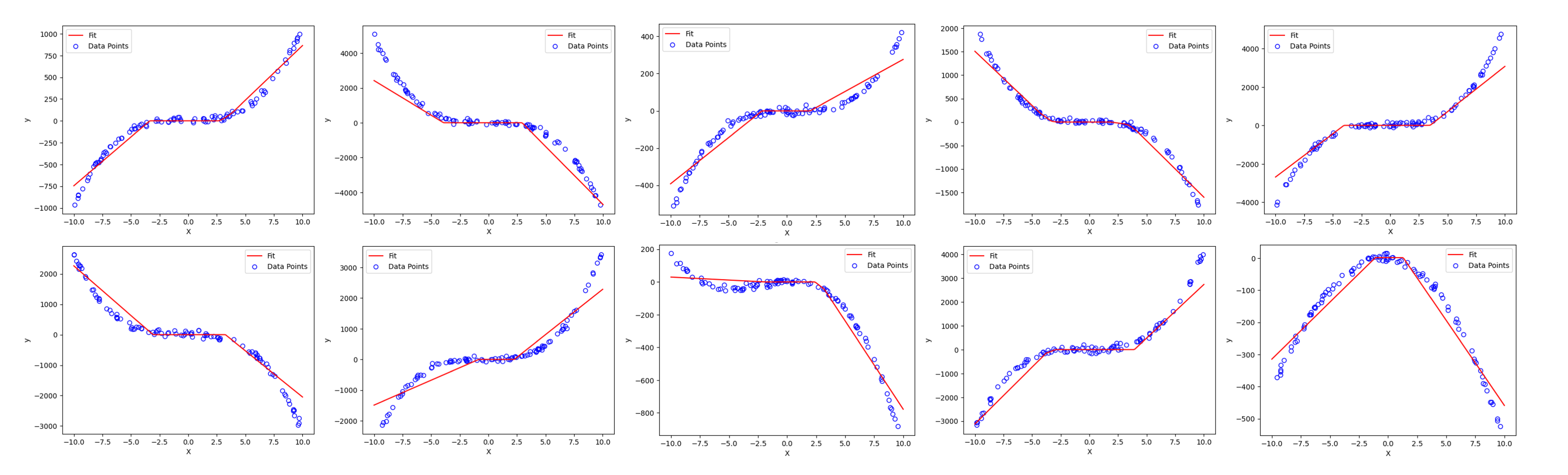}
    \caption{Testing on a regression task using the SGD optimizer with L2 regularization. 
    \textbf{R² scores:} [0.9682, 0.8859, 0.9195, 0.9683, 0.9332, 0.9335, 0.2267, 0.9505, 0.9344, 0.9429]. 
    \textbf{Norm values:} [50.65, 602.94, 12.17, 116.28, 427.72, 241.51, 1171.22, 20.47, 405.11, 10.81].}
    \label{fig:SGD_regn_regression}
\end{figure}

To further illustrate the impact of L2 regularization, we provide visual comparisons of the fitted curves obtained using different optimization settings. Figure \ref{fig:DMO_no_regn_regression}, \ref{fig:DMO_regn_regression}, \ref{fig:SGD_no_regn_regression}, \ref{fig:SGD_regn_regression} presents results on fitting regression dataset for the following cases:

\begin{enumerate}
    \item \textbf{Vanilla LO} – A learned optimizer trained without L2 regularization, later meta tested on similar dataset, resulting in larger parameter magnitudes.
    \item \textbf{LO with L2 Regularization} – A learned optimizer trained with L2 regularization during meta training, demonstrating its ability to incorporate regularization effects even during meta-testing.
    \item \textbf{Vanilla SGD} – A standard SGD optimizer without regularization, fitting the same data leading to higher-magnitude parameters.
    \item \textbf{SGD with L2 Regularization} – A standard SGD optimizer with L2 regularization, showcasing the expected reduction in norm.
\end{enumerate}

These visualizations further validate the results in Figure presented in the main paper comparing the change in $L_2$ norm of weights, reinforcing the effectiveness of learned optimizers in adapting regularization properties.

\section{Learning a regularizer}

The learned optimizer is an LSTM-based model that iteratively updates parameters of an optimizee model. It processes gradients through an LSTM and outputs parameter updates via a linear transformation. The training process incorporates initialization, preprocessing, curriculum learning, and stability-focused regularization.

\subsection{Optimizer Architecture}
The optimizer \textbf{initializes} with an optimizee model, a specified number of LSTM layers $L$, hidden size $H$, and an optional \textbf{preprocessing} mechanism. If preprocessing is enabled, the LSTM receives two-dimensional inputs; otherwise, it operates on single-dimensional inputs. The preprocessing mechanism transforms large gradients logarithmically while scaling smaller gradients exponentially.

\subsection*{Forward Pass of the Learned Optimizer}

During the \textbf{forward pass}, the learned optimizer takes as input the gradient vector $\mathbf{g}_t$ of the optimizee and processes it through a recurrent transformation $\mathcal{F}_\phi$, parameterized by the optimizer’s LSTM weights $\phi$. This transformation uses the previous hidden state $(\mathbf{h}_{t-1}, \mathbf{c}_{t-1})$ to produce an update direction $\mathbf{o}_t$:

\begin{equation}
    \mathbf{o}_t = \mathcal{F}_\phi(\mathbf{g}_t, \mathbf{h}_{t-1}, \mathbf{c}_{t-1}).
\end{equation}

The output $\mathbf{o}_t$ is then added directly to the optimizee parameters. Optional preprocessing of gradients is also applied prior to LSTM input. These steps are outlined in Algorithm \ref{algo:lo_forwardpass}.

\begin{algorithm}\label{algo:lo_forwardpass}
    \caption{Learned Optimizer Forward Pass}
    \begin{algorithmic}[1]
        \REQUIRE Gradient vector $x$, hidden state $(h_t, c_t)$, current parameters $\theta_t$
        \ENSURE Updated parameters $\theta_{t+1}$, new hidden state $(h_{t+1}, c_{t+1})$

        \IF{preprocessing is enabled}
            \STATE Initialize $z \in \mathbb{R}^{|\phi| \times 2}$
            \STATE Identify indices $i$ where $|x_i| \geq \tau$ \hfill \COMMENT{$\tau = e^{-p}$}
            \STATE $z_{i,0} \leftarrow \log(|x_i| + \epsilon)/p,\quad z_{i,1} \leftarrow \text{sign}(x_i)$
            \STATE $z_{\neg i,0} \leftarrow -1,\quad z_{\neg i,1} \leftarrow e^p \cdot x_{\neg i}$
            \STATE $x \leftarrow z$
        \ENDIF

        \STATE $(o_t, (h_{t+1}, c_{t+1})) \leftarrow \text{LSTM}(x, (h_t, c_t))$
        \STATE $\theta_{t+1} \leftarrow \theta_t + o_t$ \hfill \COMMENT{Parameter update}
        
        \RETURN $\theta_{t+1}$, $(h_{t+1}, c_{t+1})$
    \end{algorithmic}
\end{algorithm}

\subsection*{Meta-Update via Truncated Backpropagation Through Time}

To train the learned optimizer, we perform unrolled optimization over a freshly initialized optimizee model. For each outer-loop iteration, the optimizee undergoes several update steps (an unroll) using the learned optimizer, which generates parameter updates based on current gradients.

At each step \(t\) of the unroll, the optimizee model computes a loss and its gradient. This gradient \(\mathbf{g}_t\) is input to the learned optimizer, which predicts an update \(\Delta \mathbf{\phi}_t\) using its internal LSTM-based architecture. The optimizer parameters are then updated as:

\begin{equation}
    \mathbf{\phi}_{t+1} = \mathbf{\phi}_t + \Delta \mathbf{\phi}_t,
\end{equation}

where \(\Delta \mathbf{\phi}_t\) is obtained by backpropagating through the optimizee loss computed at the final step of the unroll. This serves as the meta-objective, and gradients are accumulated over multiple batches and optimizee initializations to train the optimizer via truncated backpropagation through time.

The following section outlines additional training strategies, including curriculum learning schedules, optimizer configurations, and other practical details used to stabilize and improve the meta-training process.

\section{Training Methodology for Learned Optimizers}

To progressively enhance the optimizer’s capability, we implement a curriculum learning strategy that gradually increases the number of unrolling steps during training and evaluation:

\begin{itemize}
    \item \textbf{Training Unrolling Steps} (\(\mathcal{N}_{\text{train}}\)): A sequence of increasing steps is defined as 
    \[
    \mathcal{N}_{\text{train}} = \{100, 200, 500, 1000\}.
    \]
    \item \textbf{Evaluation Unrolling Steps} (\(\mathcal{N}_{\text{eval}}\)): To assess generalization to longer horizons, evaluation steps are set as 
    \[
    \mathcal{N}_{\text{eval}} = \{200, 500, 1000\},
    \]
    intentionally excluding the initial training step.
    
    \item \textbf{Smoothing Regularization}: To ensure stability in optimizer updates, we use a perturbation-based regularization strategy. Given the optimizee state at time step \( t \), denoted as \( s_t \), we generate a perturbed state \( s'_t \) within an \( \epsilon \)-ball neighborhood $ s'_t \in B(s_t, \epsilon)$, where \( B(s_t, \epsilon) \) represents the set of all states within an \( \epsilon \)-radius under the \( l_{\infty} \) norm. The optimizer produces parameter updates \( u(s_t) \) and \( u(s'_t) \), and to encourage smoothness, we minimize their worst-case discrepancy:
    
    \begin{equation}
    R_{t+1}(\phi) = \max_{s'_t \in B(s_t, \epsilon)} \| u(s_t) - u(s'_t) \|^2.
    \end{equation}

    To efficiently compute the worst-case discrepancy \( R_{t+1}(\phi) \), we employ an iterative projected gradient ascent (PGA) approach. Specifically, we iteratively update the perturbed state \( s'_t \) within the bounded \( \epsilon \)-ball using the following procedure: $ s'_t \leftarrow s'_t + \alpha \cdot \text{sign}(\nabla_{s'_t} R_{t+1}(\phi)) $ where \( \alpha \) is the step size. To ensure that \( s'_t \) remains within the allowable perturbation bound, we apply the projection step: $ s'_t = \text{Proj}_{B(s_t, \epsilon)}(s'_t)$, where projection ensures that \( s'_t \) stays within the \( \epsilon \)-radius of \( s_t \). This iterative procedure is executed for a fixed number of steps, \( N_{\text{PGA}} \), refining the perturbation towards the worst-case deviation.

    This regularization term is added to the original objective, leading to the following training loss:
    
    \begin{equation}
    \mathcal{L}_{\text{meta}}(\phi) = l(f(\theta_t(\phi); x), y) + \lambda R_t(\phi).
    \end{equation}

    This formulation ensures that the optimizer produces stable parameter updates, reducing sensitivity to small variations in the optimizee’s state and improving generalization. The following steps will give rise to the total loss that include loss due to regularization.

    \item \textbf{Regularization Techniques}: To further enhance stability and generalization, we employ advanced regularization methods: 

    \begin{itemize}
        \item \textbf{Sharpness-Aware Minimization (SAM)}: SAM improves generalization by minimizing both the loss function and its sensitivity to weight perturbations. The optimization objective is  
    
            \begin{equation}
            \min_{w} \mathcal{L}_{SAM} (w), \quad \text{where} \quad \mathcal{L}_{SAM} (w) = \max_{\|\epsilon\|_2 \leq \rho} \mathcal{L} (w + \epsilon).
            \end{equation}
            
            The perturbation is approximated as  
            
            \begin{equation}
            \hat{\epsilon} = \rho \frac{\nabla_w \mathcal{L}_{\text{meta}} (w)}{\| \nabla_w \mathcal{L}_{\text{meta}} (w) \|_2},
            \end{equation}
            
            leading to the perturbed weight update:  
            
            \begin{equation}
            \phi^+ = \phi + \hat{\epsilon}, \quad \phi \leftarrow \phi^+ - \eta \nabla_\phi \mathcal{L}_{meta} (\phi^+).
            \end{equation}

        \item \textbf{Surrogate Gap Guided Sharpness-Aware Minimization (GSAM)}: GSAM extends SAM by introducing a surrogate gap \( h(\phi) \approx \mathcal{L}_{\text{meta}}(\phi_{\text{adv}}) - \mathcal{L}_{\text{meta}}(\phi) \), which better captures sharpness.  SAM minimizes a perturbed loss defined as  

            \begin{equation}
            \mathcal{L}_{\text{meta}, p}(\phi) = \max_{\|\delta\| \leq \rho} \mathcal{L}_{\text{meta}}(\phi + \delta),
            \end{equation}

            which represents the worst-case loss within a neighborhood of radius \( \rho \) around \( \phi \). GSAM improves upon this by adjusting the perturbation to better align with the sharpness objective. The perturbation is computed as,
            
            \begin{equation}
            \Delta \phi_t = \rho_t \frac{\nabla \mathcal{L}_{\text{meta}}^{(t)}}{\|\nabla \mathcal{L}_{\text{meta}}^{(t)}\| + \epsilon},
            \end{equation}  
            
            leading to the adversarial weight \( \phi_{\text{adv}} = \phi_t + \Delta \phi_t \). GSAM follows two steps: (1) a descent step on the perturbed loss, similar to SAM, and (2) gradient decomposition into parallel and orthogonal components relative to \( \nabla \mathcal{L}_{\text{meta}}(\phi_{\text{adv}}) \), performing an ascent step along the orthogonal component to minimize the surrogate gap. The final update is  
            
            \begin{equation}
            \phi_{t+1} = \phi_t - \eta_t (\nabla \mathcal{L}_{\text{meta}, p}^{(t)} - \alpha \nabla_{\perp} \mathcal{L}_{\text{meta}}^{(t)}).
            \end{equation}

            \item \textbf{Gradient-norm Aware Minimization (GAM)}: GAM refines generalization by incorporating first-order sharpness into the loss optimization process. The sharpness-aware regularization is given by:  

                \begin{equation}
                \mathcal{L}_{\text{GAM}}(\phi) \triangleq \rho \cdot \max_{\phi' \in B(\phi, \rho)} \left\| \nabla \mathcal{L}_{\text{meta}}(\phi') \right\|.
                \end{equation}
                
                The sharpness gradient is approximated as:  
                
                \begin{equation}
                \nabla \mathcal{L}_{\text{GAM}}(\phi) \approx \rho \cdot \nabla \left\| \nabla \mathcal{L}_{\text{meta}}(\phi_{\text{adv}}) \right\|, 
                \quad \phi_{\text{adv}} = \phi + \rho \cdot \frac{f}{\| f \|},
                \end{equation}
                
                with \( f = \nabla \left\| \nabla \mathcal{L}_{\text{meta}}(\phi) \right\| \). The gradient term simplifies using Hessian-vector products:
                
                \begin{equation}
                \nabla \left\| \nabla \mathcal{L}_{\text{meta}}(\phi) \right\| = \frac{\nabla^2 \mathcal{L}_{\text{meta}}(\phi) \cdot \nabla \mathcal{L}_{\text{meta}}(\phi)}{\left\| \nabla \mathcal{L}_{\text{meta}}(\phi) \right\|}.
                \end{equation}

        \item \textbf{Compute the Total Loss}: The total loss consists of the primary task loss \( \mathcal{L}_{\text{task}} \), the smoothing regularization term \( \mathcal{L}_{\text{smooth}} \), and a selected sharpness-aware loss from \{SAM, GSAM, GAM\}:

            \begin{equation}
            \mathcal{L}(\phi) = \mathcal{L}_{\text{task}}(\phi) + \lambda_{\text{smooth}} \mathcal{L}_{\text{smooth}}(\phi) + \lambda_{\text{reg}} \mathcal{L}_{\text{reg}}(\phi),
            \end{equation}
            
            where \( \mathcal{L}_{\text{reg}}(\phi) \) corresponds to either \( \mathcal{L}_{SAM} \), \( \mathcal{L}_{GSAM} \), or \( \mathcal{L}_{GAM} \), depending on the chosen regularization strategy.
            
            The objective is to find the optimal optimizer parameters \( \phi^* \) that minimize the total loss:
            
            \begin{equation}
            \phi^* = \arg\min_{\phi} \mathcal{L}(\phi).
            \end{equation}

            \item \textbf{Update the Optimizer Parameters}: The gradient of the total loss with respect to the optimizer parameters \( \phi \), $\nabla_{\phi} \mathcal{L}(\phi)$ is calculated. The optimizer parameters \( \phi \) are updated using a base optimizer such as SGD or Adam. For GSAM and GAM, additional scheduling mechanisms are employed to adaptively adjust learning rates and regularization parameters during training:
            
            \begin{itemize}
                \item For \textbf{GAM}, a cosine annealing learning rate scheduler is applied alongside proportion-based schedulers for gradient sharpness parameters.
                \item For \textbf{GSAM}, a cosine-based learning rate scheduler and a proportion-based \( \rho \)-scheduler are used to regulate the sharpness adjustment.
            \end{itemize}

    \end{itemize}

\end{itemize}

\section{Measuring Sharpness Using Other Evaluation Metrics}

In Section Regularization in the main paper, we examined optimization mechanisms employing various regularizers. We also discussed learning a regularizer using the Meta-Training methodology. 

We trained a LO using the regularized loss, anticipating it would internalize the general mechanism of seeking minima with properties characteristic of SAM, GAM, or GSAM—specifically, maximum loss in the neighborhood, maximum gradient norm in the neighborhood, or Surrogate gap at PoC, respectively. To assess the satisfaction of these properties in the vicinity of a point, we:

\begin{itemize}
    \item Conducted a 10-step Projected Gradient Ascent (PGA) on:
    \begin{itemize}
        \item The difference between the loss at a point and the loss at the PoC (for \textbf{SAM} and \textbf{GSAM}).
        \item The difference between the gradient norm at a perturbed point and the gradient norm at the PoC (for \textbf{GAM}).
    \end{itemize}
    \item Varied the size of the neighborhood in which we sought the maximum value of the loss or gradient norm. (Neighborhood ball sizes: [0.001, 0.005, 0.01, 0.05, 0.1]).
    \item Evaluated the property at two different training iterations: one at convergence (determined by a stopping criterion) and another after a predetermined number of training iterations.
\end{itemize}

\subsection{Projected Gradient Ascent (PGA)}

Projected Gradient Ascent (PGA) is an optimization technique aimed at maximizing a function \( f(x) \) while ensuring that the parameter vector \( x \) remains within a specified feasible set \( C \). The method involves iteratively updating \( x \) by moving in the direction of the gradient of \( f \) and then projecting the result back onto \( C \) to maintain feasibility.

The update rule at iteration \( k \) is given by:

\[
x_{k+1} = P_C(x_k + \alpha_k \nabla f(x_k)),
\]

where \( \alpha_k \) is the step size, \( \nabla f(x_k) \) is the gradient of \( f \) at \( x_k \), and \( P_C(\cdot) \) denotes the projection operator onto the set \( C \).

In scenarios where \( C \) is defined as a norm ball centered at an initial point \( x_0 \) with radius \( \rho \), the projection operator ensures that the updated parameters do not deviate beyond \( \rho \) from \( x_0 \). The projection function can be expressed as:

\[
P_C(y) = x_0 + \rho \frac{y - x_0}{\|y - x_0\|}.
\]

The update function is given by:

\[
x_{k+1} = \text{proj}_{B(x_0, \rho)}(x_k + \alpha \nabla_x f(x_k))
\]

where \( k \) denotes the current iteration of PGA, \( \alpha \) is the step size, and \( f \) is the considered function.

For instance, considering the loss as the functional value, the value obtained in the final (10th) step provides an estimate of the maximum value of loss within that neighborhood. For clarity, we focus on the value of the property in the last step. We will discuss the various factors involved in our case in detail in the next section.

\subsection{PGA for Maximum Loss in a neighbourhood and Surrogate Gap Estimation}

We employ Projected Gradient Ascent (PGA) to determine the maximum validation loss within a constrained neighborhood of the model parameters. The key configuration parameters are:

\begin{itemize}
    \item \textbf{Perturbation Radius ($\epsilon$)}: Bounds the maximum allowable deviation per parameter within an $L_\infty$ ball centered at the original model.
    \item \textbf{Step Size ($\delta$)}: Set as $\epsilon/10$, it controls the magnitude of each update during PGA to ensure gradual, stable progress.
    \item \textbf{Perturbation Norm and Projection}: Perturbations follow the $L_\infty$ norm constraint, enforced via element-wise clipping after each update to stay within $[\theta_i - \epsilon, \theta_i + \epsilon]$.
    \item \textbf{Initialization}: Parameters are perturbed with Gaussian noise to explore diverse regions in the local neighborhood.
    \item \textbf{Iterations}: The update loop runs for 10 steps, progressively maximizing the validation loss within the norm-bound region.
    \item \textbf{Loss Maximization}: At each step, the validation loss is evaluated and maximized by ascending the gradient of the loss difference relative to the base model.
    \item \textbf{Gradient Computation}: Gradients are taken with respect to the change in loss (perturbed minus baseline), targeting the direction of greatest performance degradation.
    \item \textbf{Parameter Flattening}: Model parameters are flattened to a single vector for unified gradient updates and norm enforcement.
    \item \textbf{Loss Aggregation}: The loss is averaged over the validation dataset to reflect global performance under perturbation.
    \item \textbf{Surrogate Gap Estimate}: The difference between perturbed and baseline loss quantifies the worst-case degradation and serves as a surrogate robustness measure.
\end{itemize}

This approach provides a systematic framework for robustness assessment by identifying the most vulnerable parameter configurations within a controlled perturbation space, thereby enabling precise characterization of model stability under parameter variations.

\begin{algorithm}
    \caption{Maximum Loss via Projected Gradient Descent with Surrogate Gap}
    \begin{algorithmic}[1]
        \REQUIRE Model parameters $\theta$, perturbation radius $\epsilon$, validation data $\mathcal{D}$
        \ENSURE Perturbed parameters $\theta^*$ with maximum loss, surrogate gap estimate
        
        \STATE $\ell_{\text{base}} \leftarrow \frac{1}{|\mathcal{D}|}\sum_{(x,y) \in \mathcal{D}} \mathcal{L}(f_{\theta}(x), y)$ \hfill \textit{// baseline loss}
        \STATE $\theta_0 \leftarrow \theta + \xi$ where $\xi \sim \mathcal{N}(0, I)$ \hfill \textit{// random initialization}
        \STATE $\alpha \leftarrow \epsilon/10$ \hfill \textit{// step size}
        
        \FOR{$t = 0, 1, \ldots, 9$}
            \STATE $\ell_t \leftarrow \frac{1}{|\mathcal{D}|}\sum_{(x,y) \in \mathcal{D}} \mathcal{L}(f_{\theta_t}(x), y)$ \hfill \textit{// current loss}
            \STATE $\Delta \leftarrow \ell_t - \ell_{\text{base}}$ \hfill \textit{// loss difference}
            \STATE $g \leftarrow \nabla_{\theta_t} \Delta$ \hfill \textit{// gradient to maximize loss}
            \STATE $\theta_{t+1} \leftarrow \theta_t + \alpha \cdot \text{sign}(g)$ \hfill \textit{// gradient ascent}
            \STATE $\theta_{t+1} \leftarrow \text{clip}(\theta_{t+1}, \theta - \epsilon \mathbf{1}, \theta + \epsilon \mathbf{1})$ \hfill \textit{// project to $L_\infty$ ball}
        \ENDFOR
        
        \STATE $\ell_{\text{final}} \leftarrow \frac{1}{|\mathcal{D}|}\sum_{(x,y) \in \mathcal{D}} \mathcal{L}(f_{\theta_{10}}(x), y)$ \hfill \textit{// final loss}
        \STATE $\text{gap} \leftarrow \ell_{\text{final}} - \ell_{\text{base}}$ \hfill \textit{// surrogate gap estimate}
        \RETURN $\theta_{10}, \ell_{\text{final}}, \text{gap}$
    \end{algorithmic}
\end{algorithm}

\subsection{PGA for Maximum Norm in a Neighbourhood}

We employ Projected Gradient Ascent (PGA) to iteratively perturb the model parameters, aiming to maximize the gradient norm of the loss function within a constrained neighborhood. The configuration parameters, including perturbation radius, step size, norm constraint ($L_\infty$), random initialization, and iteration count, follow the same setup as described in the previous section.

\begin{itemize}
    \item \textbf{Gradient Norm Maximization}: In each iteration, the loss is computed over the dataset, followed by evaluation of the gradient with respect to model parameters. The gradient norm serves as a surrogate measure of sensitivity. Updates are computed to maximize the increase in this norm, with each perturbation projected back to the $\epsilon$-ball around the original parameters.

\end{itemize}

\begin{algorithm}[H]\label{algo:pgd_for_max_grad_norm}
    \caption{PGD to Maximize Gradient Norm ($L_\infty$ Norm)}
    \begin{algorithmic}[1]
        \REQUIRE Model parameters $\theta$, perturbation radius $\epsilon$, validation data $\mathcal{D}$
        \ENSURE Final gradient norm $g_{\text{final}}$

        \STATE Compute base gradient norm: $g_{\text{base}} \leftarrow \left\lVert \nabla_\theta \mathcal{L}(f_\theta(\mathcal{D})) \right\rVert$
        \STATE Initialize $\theta_0 \leftarrow \theta + \xi$, where $\xi \sim \mathcal{N}(0, I)$
        \STATE $\alpha \leftarrow \epsilon$

        \FOR{$t = 0, 1, \ldots, 9$}
            \STATE $g_t \leftarrow \left\lVert \nabla_{\theta_t} \mathcal{L}(f_{\theta_t}(\mathcal{D})) \right\rVert$
            \STATE $\Delta \leftarrow g_t - g_{\text{base}}$
            \STATE $g \leftarrow \nabla_{\theta_t} \Delta$
            \STATE $\theta_{t+1} \leftarrow \theta_t + \alpha \cdot \text{sign}(g)$
            \STATE $\theta_{t+1} \leftarrow \text{clip}(\theta_{t+1}, \theta - \epsilon, \theta + \epsilon)$
        \ENDFOR

        \STATE $g_{\text{final}} \leftarrow \left\lVert \nabla_{\theta_{10}} \mathcal{L}(f_{\theta_{10}}(\mathcal{D})) \right\rVert$
        \RETURN $g_{\text{final}}$
    \end{algorithmic}
\end{algorithm}

\section{Meta-Test Results: LO Capturing Regularization Properties Across Architectures, Datasets, and Optimization Tasks}

\label{appendix:results_samgamgsam}

This section expands upon the meta-testing evaluation of the learned optimizer (LO) by analyzing its behavior across a range of architectures, datasets, and regularization strategies. Moving beyond the results presented in the main paper, where a fixed neighborhood radius of $0.01$ was examined, we now present an in-depth study using varied radii—specifically $0.001$, $0.005$, $0.05$, and $0.1$—to assess the optimizer's sensitivity and generalization capacity. The analysis highlights how the LO internalizes characteristic properties of the regularizers it was exposed to during training, and reflects those properties when guiding optimization trajectories in unseen tasks. We evaluate this effect in the context of SAM, GSAM, and similar approaches using classification benchmarks.

\subsection{MLP with Varying Architectures and Regularization Techniques}

\textbf{Task:} MNIST classification using a MLP with a single hidden layer comprising 20 neurons and sigmoid activation functions.

\begin{figure}[htbp]
    \centering
    \includegraphics[width=0.95\linewidth]{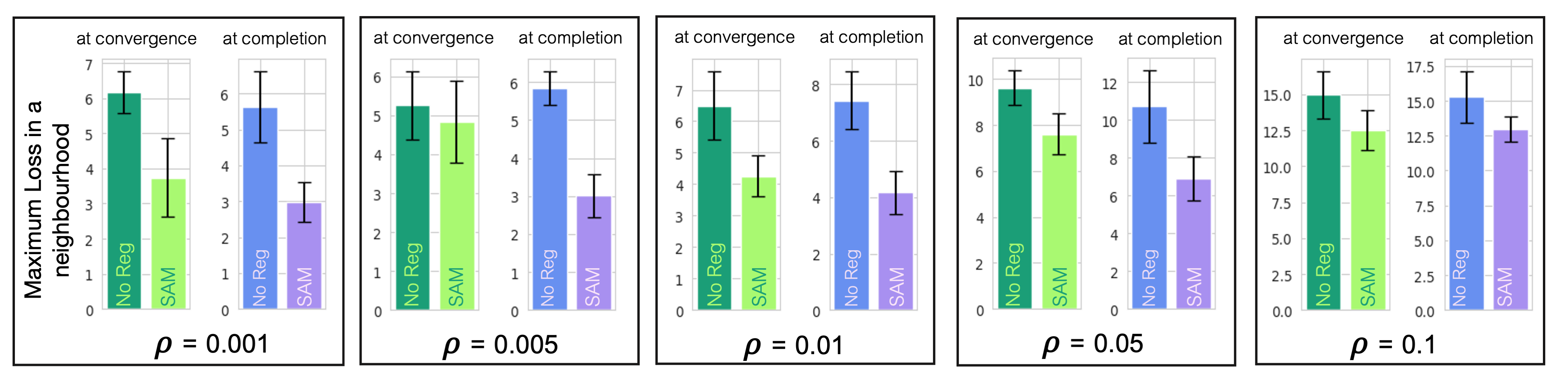}
    \caption{Meta-Test Testing Results for \textbf{MLP} on \textbf{MNIST} dataset with \textbf{SAM} regularization.}
    \label{fig:mlp_mnist_sam}
\end{figure}

\begin{figure}[htbp]
    \centering
    \includegraphics[width=0.95\linewidth]{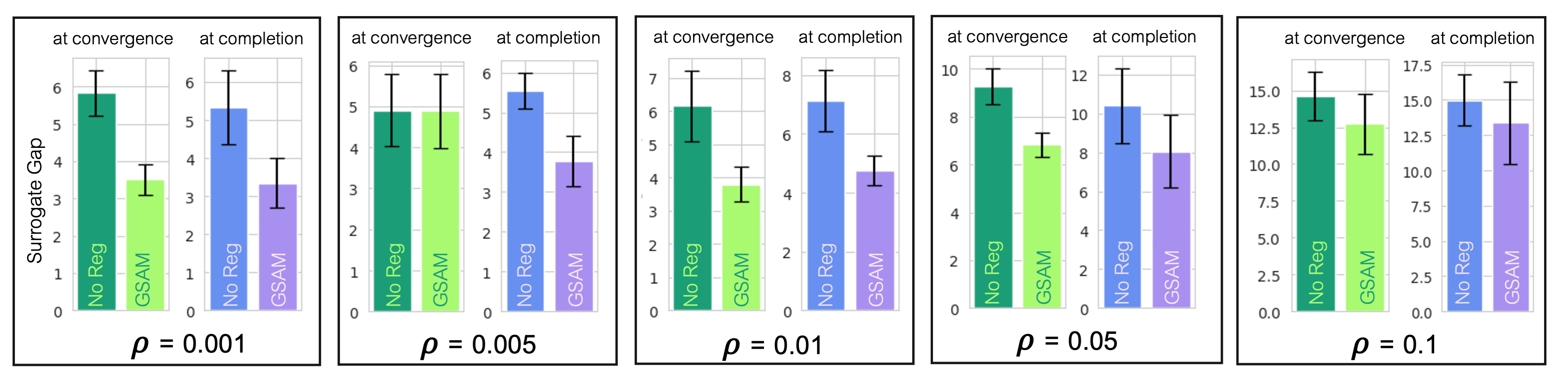}
    \caption{Meta-Test Testing Results for \textbf{MLP} on \textbf{MNIST} dataset with \textbf{GSAM} regularization.}
    \label{fig:mlp_mnist_gsam}
\end{figure}

\begin{figure}[htbp]
    \centering
    \includegraphics[width=0.95\linewidth]{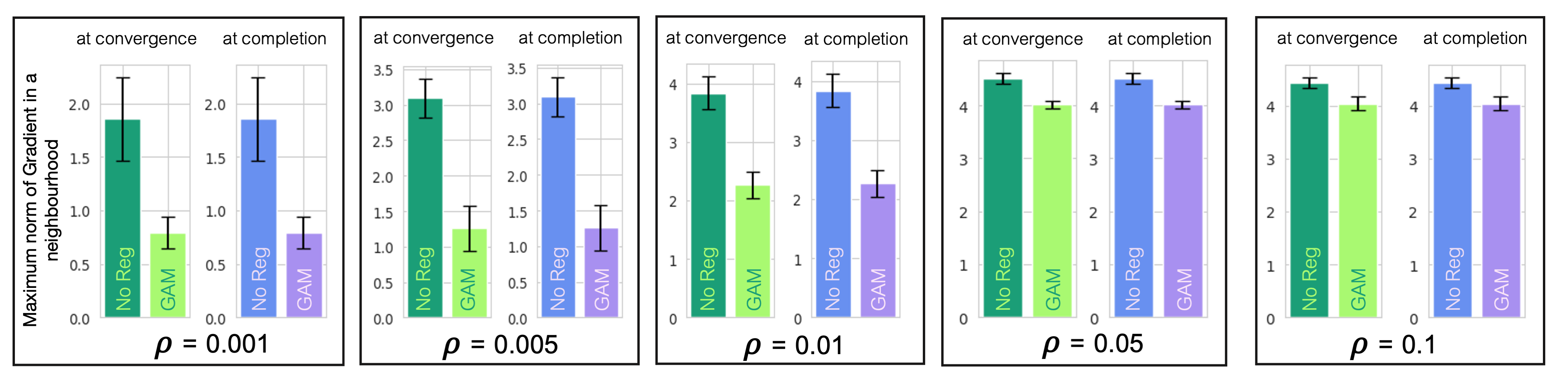}
    \caption{Meta-Test Testing Results for \textbf{MLP} on \textbf{MNIST} dataset with \textbf{GAM} regularization.}
    \label{fig:mlp_mnist_gam}
\end{figure}

\textbf{TASK} : FMNIST Classification, using MLP (Single Layer of 20 neurons) with
sigmoid activation function.

\begin{figure}[htbp]
    \centering
    \includegraphics[width=0.95\linewidth]{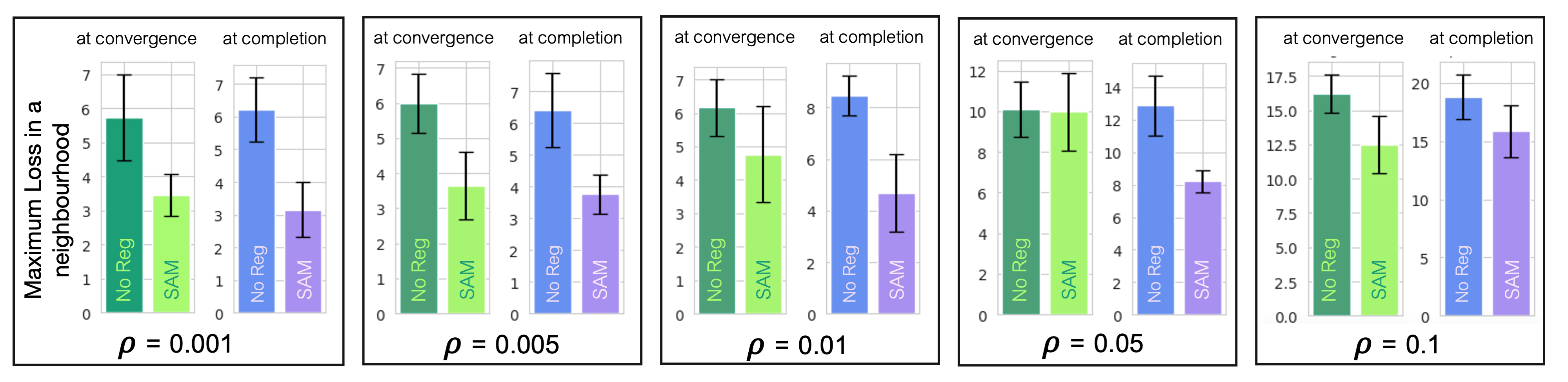}
    \caption{Meta-Test Testing Results for \textbf{MLP} on \textbf{FMNIST} dataset with \textbf{SAM} regularization.}
    \label{fig:mlp_fmnist_sam}
\end{figure}

\begin{figure}[htbp]
    \centering
    \includegraphics[width=0.95\linewidth]{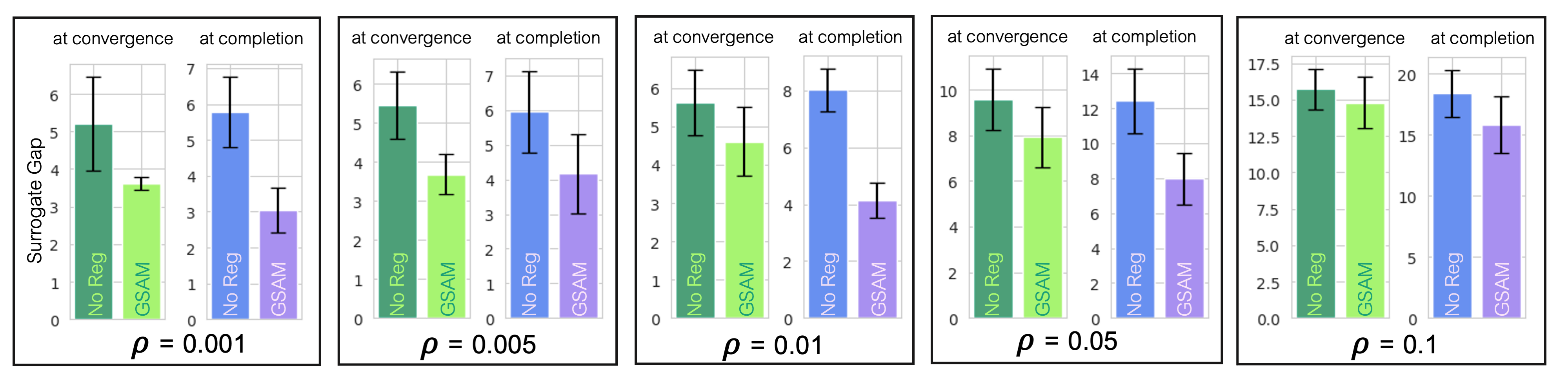}
    \caption{Meta-Test Testing Results for \textbf{MLP} on \textbf{FMNIST} dataset with \textbf{GSAM} regularization.}
    \label{fig:mlp_fmnist_gsam}
\end{figure}

\begin{figure}[htbp]
    \centering
    \includegraphics[width=0.95\linewidth]{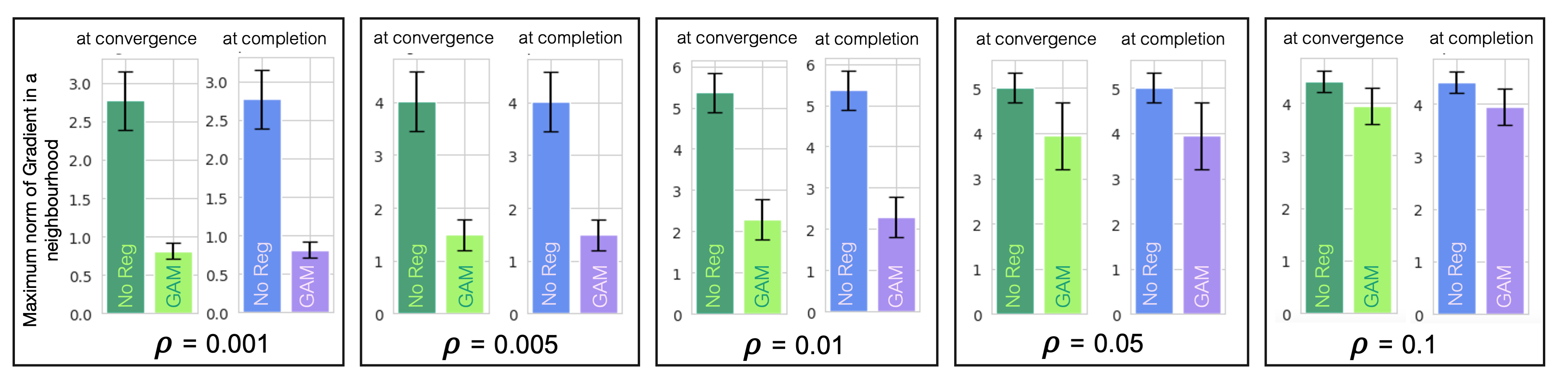}
    \caption{Meta-Test Testing Results for \textbf{MLP} on \textbf{FMNIST} dataset with \textbf{GAM} regularization.}
    \label{fig:mlp_fmnist_gam}
\end{figure}

\textbf{TASK:} CIFAR10 Classification, using MLP (Single Layer of 20 neurons) with
sigmoid activation function.

\begin{figure}[htbp]
    \centering
    \includegraphics[width=0.95\linewidth]{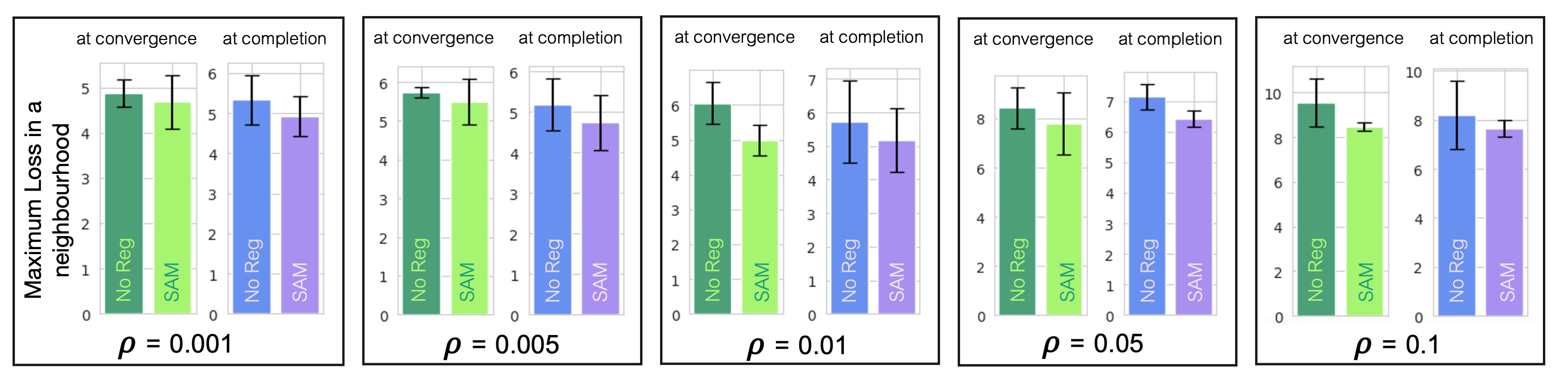}
    \caption{Meta-Test Testing Results for \textbf{MLP} on \textbf{CIFAR10} dataset with \textbf{SAM} regularization.}
    \label{fig:mlp_cifar10_sam}
\end{figure}

\begin{figure}[htbp]
    \centering
    \includegraphics[width=0.95\linewidth]{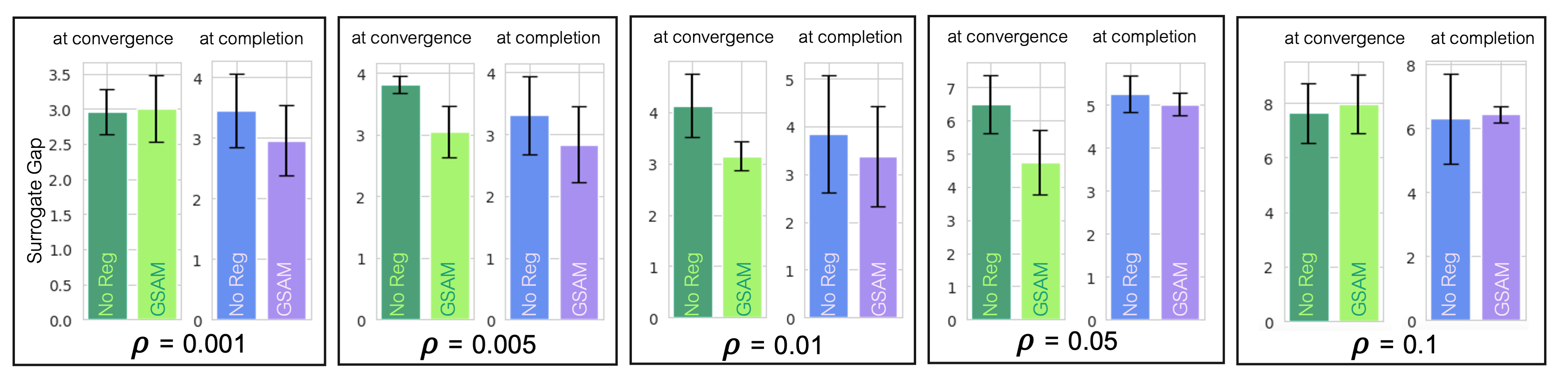}
    \caption{Meta-Test Testing Results for \textbf{MLP} on \textbf{CIFAR10} dataset with \textbf{GSAM} regularization.}
    \label{fig:mlp_cifar10_gsam}
\end{figure}

\begin{figure}[htbp]
    \centering
    \includegraphics[width=0.95\linewidth]{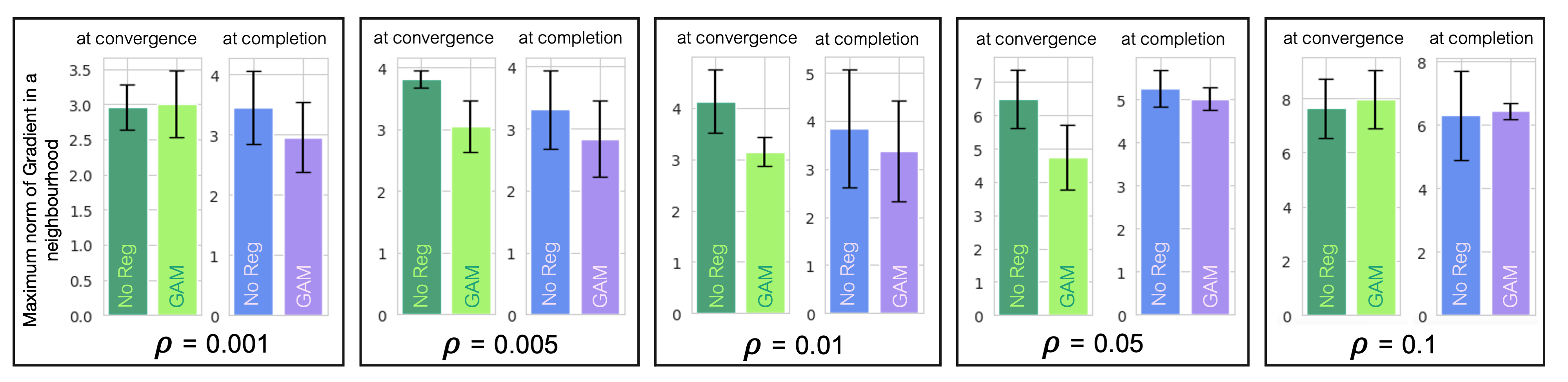}
    \caption{Meta-Test Testing Results for \textbf{MLP} on \textbf{CIFAR10} dataset with \textbf{GAM} regularization.}
    \label{fig:mlp_cifar10_gam}
\end{figure}

\subsection{MLP (with ReLU Activation Function) with Different Architectures and Different Regularization Techniques}

\textbf{TASK} : MNIST Classification, using MLP (Single Layer of 20 neurons) with
Relu activation function.

\begin{figure}[htbp]
    \centering
    \includegraphics[width=0.95\linewidth]{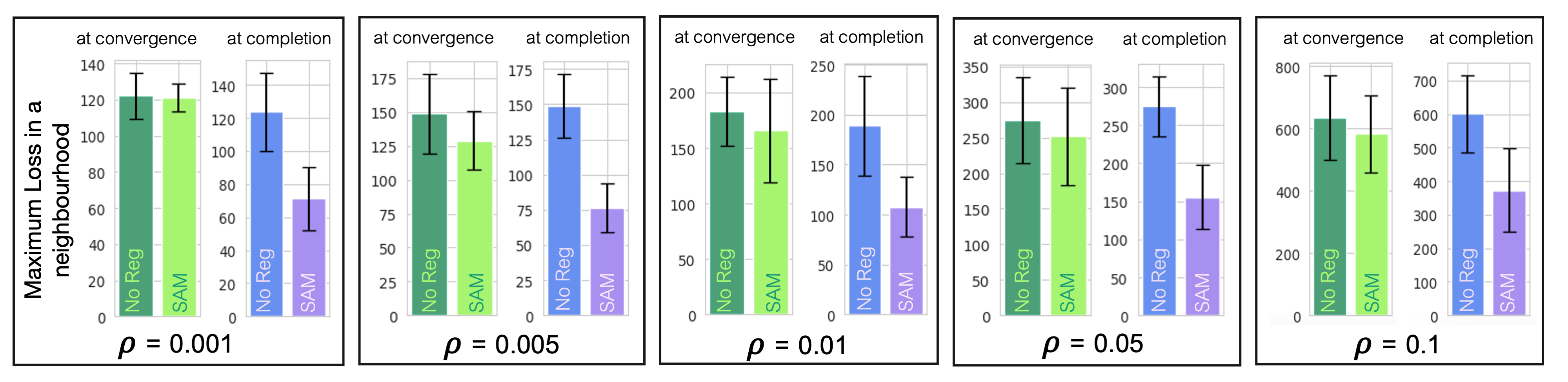}
    \caption{Meta-Test Testing Results for \textbf{MLP (ReLU)} on \textbf{MNIST} dataset with \textbf{SAM} regularization.}
    \label{fig:mlpRelu_mnist_sam}
\end{figure}

\begin{figure}[htbp]
    \centering
    \includegraphics[width=0.95\linewidth]{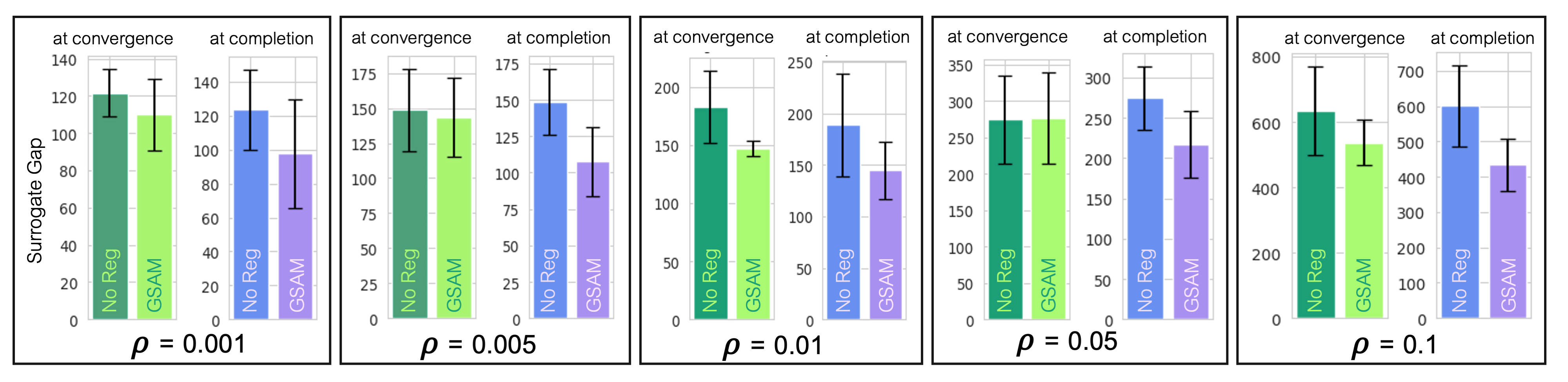}
    \caption{Meta-Test Testing Results for \textbf{MLP (ReLU)} on \textbf{MNIST} dataset with \textbf{GSAM} regularization.}
    \label{fig:mlpRelu_mnist_gsam}
\end{figure}

\begin{figure}[htbp]
    \centering
    \includegraphics[width=0.95\linewidth]{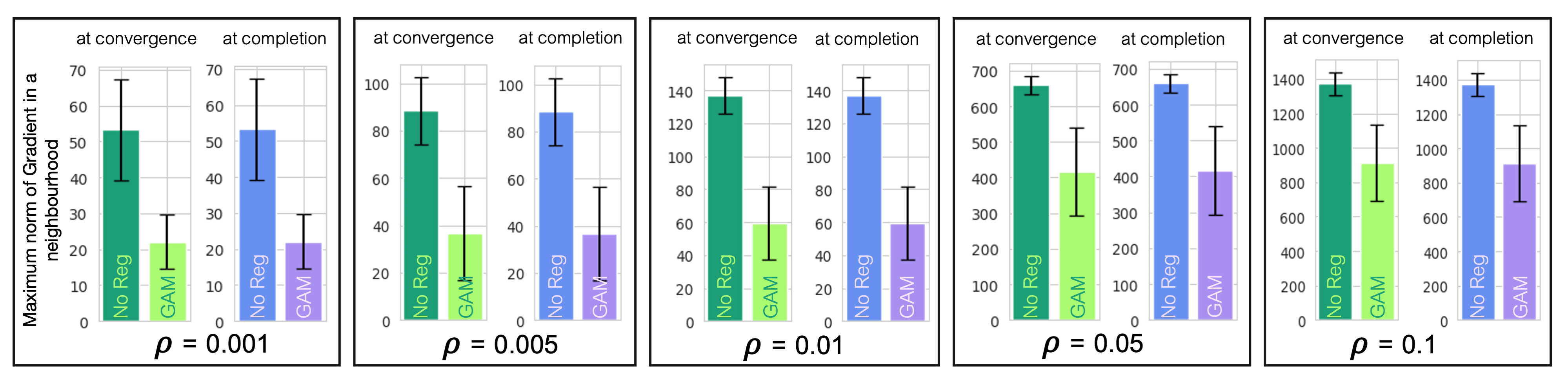}
    \caption{Meta-Test Testing Results for \textbf{MLP (ReLU)} on \textbf{MNIST} dataset with \textbf{GAM} regularization.}
    \label{fig:mlpRelu_mnist_gam}
\end{figure}

\textbf{TASK} : FMNIST Classification, using MLP (Single Layer of 20 neurons) with
Relu activation function.

\begin{figure}[htbp]
    \centering
    \includegraphics[width=0.95\linewidth]{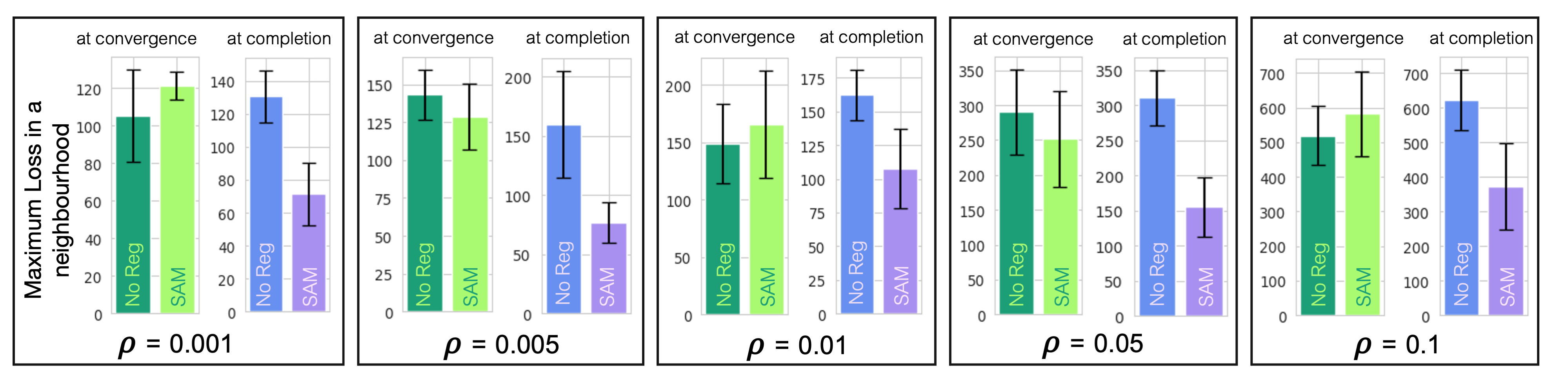}
    \caption{Meta-Test Testing Results for \textbf{MLP (ReLU)} on \textbf{FMNIST} dataset with \textbf{SAM} regularization.}
    \label{fig:mlpRelu_fmnist_sam}
\end{figure}

\begin{figure}[htbp]
    \centering
    \includegraphics[width=0.95\linewidth]{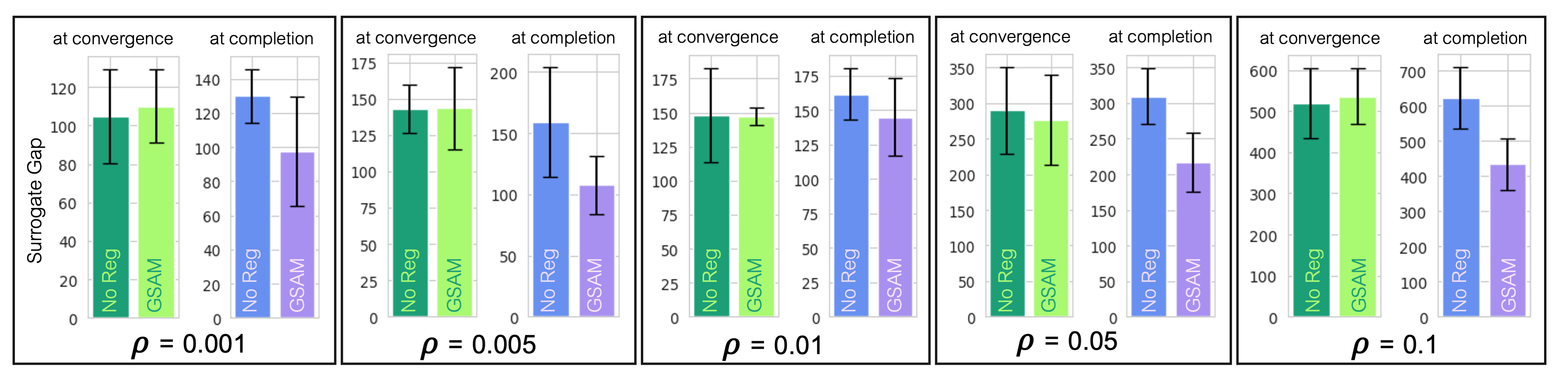}
    \caption{Meta-Test Testing Results for \textbf{MLP (ReLU)} on \textbf{FMNIST} dataset with \textbf{GSAM} regularization.}
    \label{fig:mlpRelu_fmnist_gsam}
\end{figure}

\begin{figure}[htbp]
    \centering
    \includegraphics[width=0.95\linewidth]{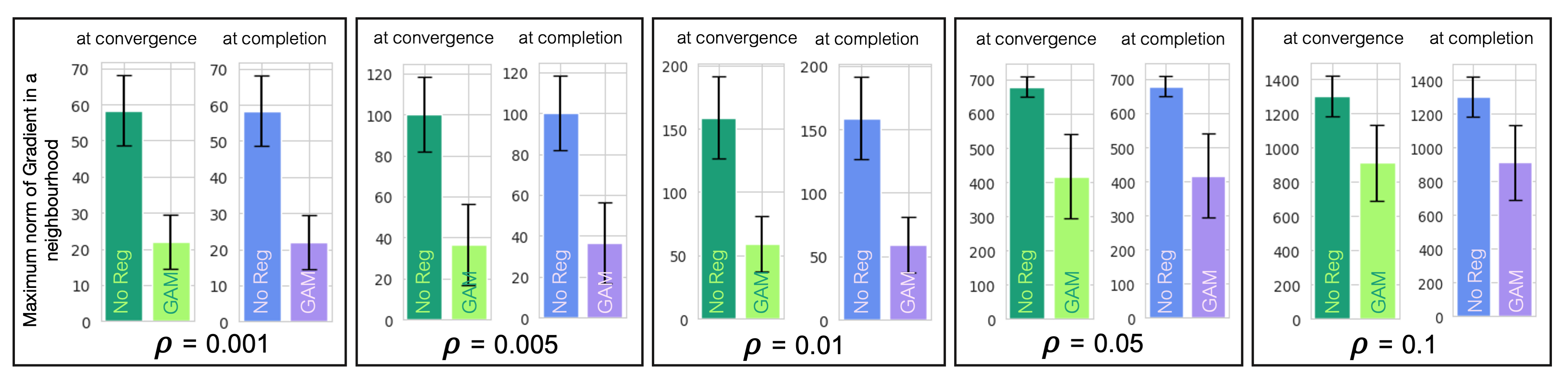}
    \caption{Meta-Test Testing Results for \textbf{MLP (ReLU)} on \textbf{FMNIST} dataset with \textbf{GAM} regularization.}
    \label{fig:mlpRelu_fmnist_gam}
\end{figure}

\textbf{TASK} : CIFAR10 Classification, using MLP (Single Layer of 20 neurons) with
Relu activation function.
\begin{figure}[htbp]
    \centering
    \includegraphics[width=0.95\linewidth]{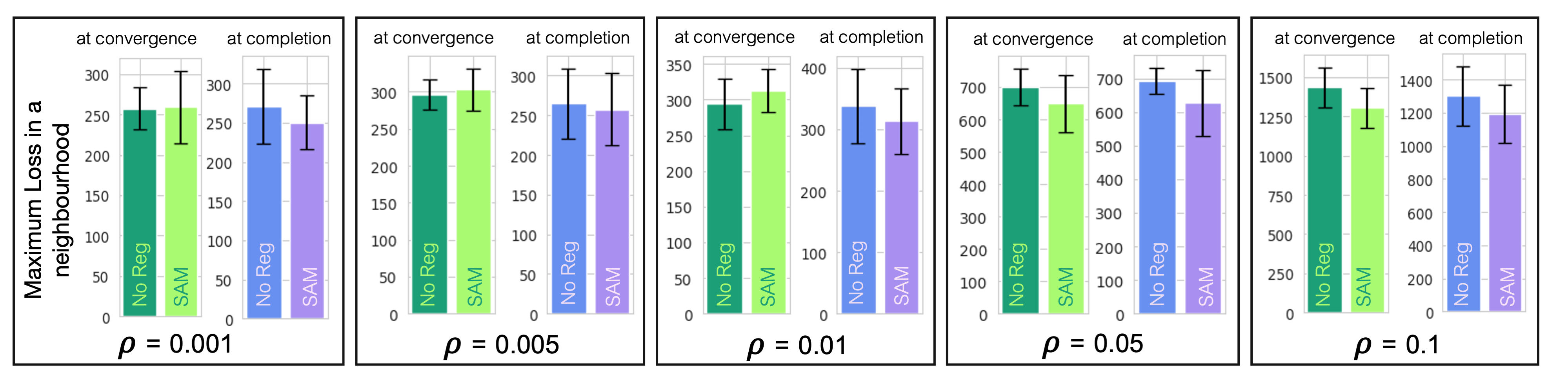}
    \caption{Meta-Test Testing Results for \textbf{MLP (ReLU)} on \textbf{CIFAR10} dataset with \textbf{SAM} regularization.}
    \label{fig:mlpRelu_cifar10_sam}
\end{figure}

\begin{figure}[htbp]
    \centering
    \includegraphics[width=0.95\linewidth]{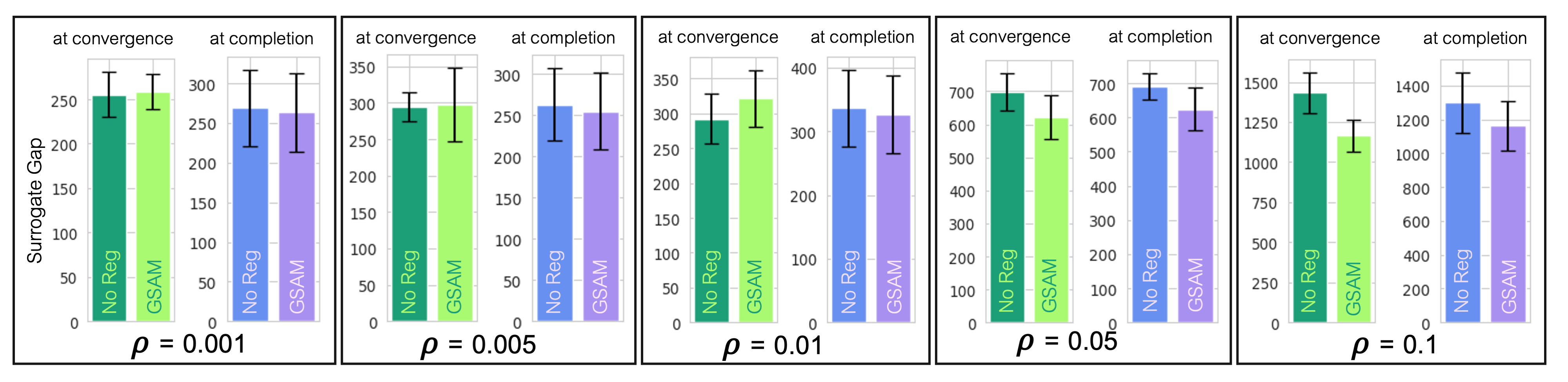}
    \caption{Meta-Test Testing Results for \textbf{MLP (ReLU)} on \textbf{CIFAR10} dataset with \textbf{GSAM} regularization.}
    \label{fig:mlpRelu_cifar10_gsam}
\end{figure}

\begin{figure}[htbp]
    \centering
    \includegraphics[width=0.95\linewidth]{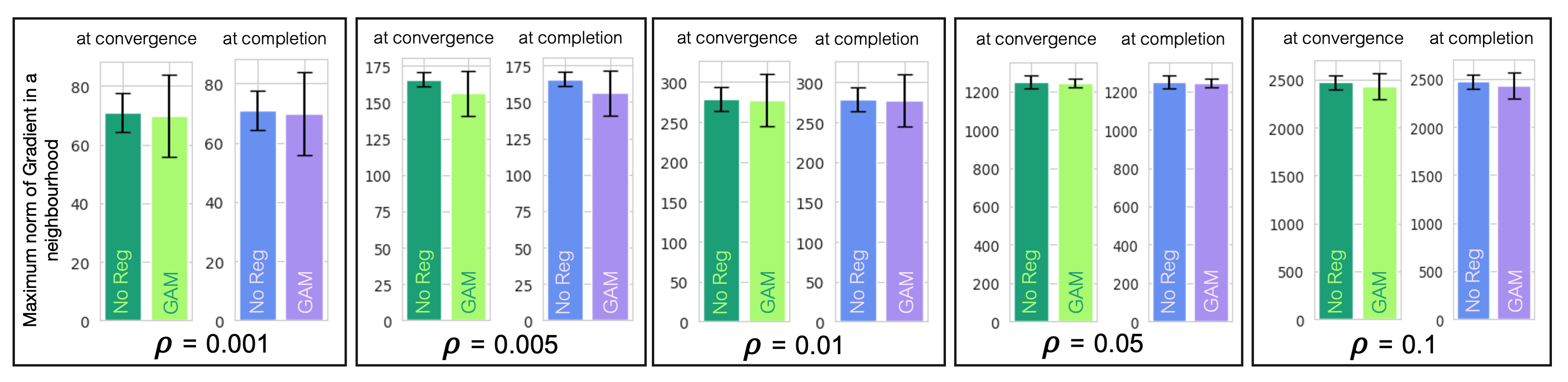}
    \caption{Meta-Test Testing Results for \textbf{MLP (ReLU)} on \textbf{CIFAR10} dataset with \textbf{GAM} regularization.}
    \label{fig:mlpRelu_cifar10_gam}
\end{figure}

\textbf{Note:} An important observation on the loss surface of the MLP-ReLU model is the notably high maximum loss encountered within a small neighborhood around the Point of Convergence (PoC). This aligns with challenges reported in prior studies~\cite{L2LGDGD, hierarchyRNN}, which discussed the difficulty of locating suitable minima for this task. Our current findings provide a hint on the same issue by highlighting the sharp variations in the local loss landscape.

\subsection{CNN with Different Architectures and Different Regularization Techniques}

\textbf{TASK} : MNIST Classification, using Convolution Neural Network (two 2D Convolution Layer with 16 3×3 kernels and 32 5×5 kernels with 2D Max Pool layer in between with kernel size 2 × 2).

\begin{figure}[htbp]
    \centering
    \includegraphics[width=0.95\linewidth]{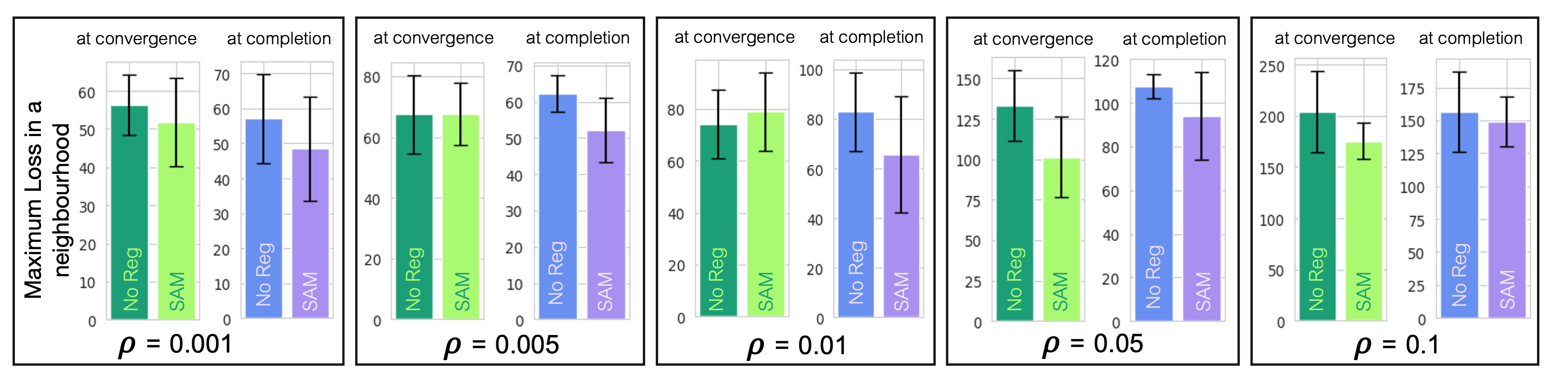}
    \caption{Meta-Test Testing Results for \textbf{CNN} on \textbf{MNIST} dataset with \textbf{SAM} regularization.}
    \label{fig:conv_mnist_sam}
\end{figure}

\begin{figure}[htbp]
    \centering
    \includegraphics[width=0.95\linewidth]{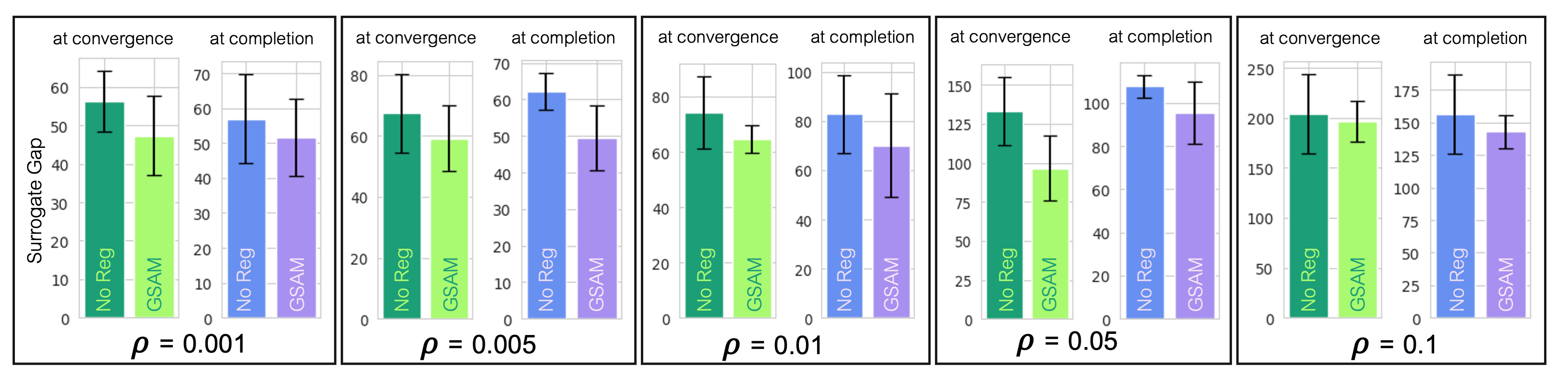}
    \caption{Meta-Test Testing Results for \textbf{CNN} on \textbf{MNIST} dataset with \textbf{GSAM} regularization.}
    \label{fig:conv_mnist_gsam}
\end{figure}

\begin{figure}[htbp]
    \centering
    \includegraphics[width=0.95\linewidth]{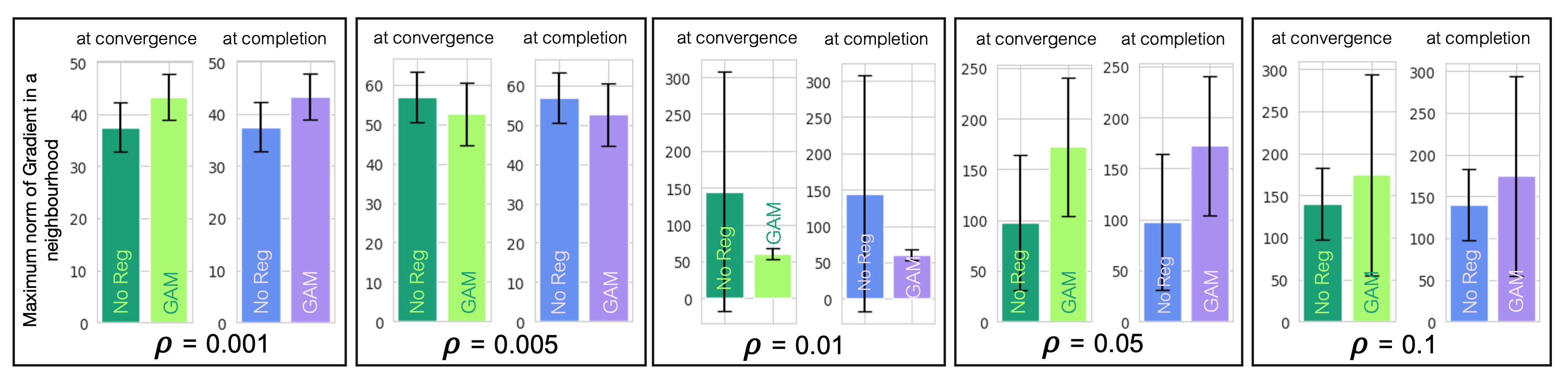}
    \caption{Meta-Test Testing Results for \textbf{CNN} on \textbf{MNIST} dataset with \textbf{GAM} regularization.}
    \label{fig:conv_mnist_gam}
\end{figure}

\textbf{TASK} : FMNIST Classification, using Convolution Neural Network (two 2D Convolution Layer with 16 3 × 3 kernels and 32 5 × 5 kernels with 2D Max Pool layer in between with kernel size 2 × 2)

\begin{figure}[htbp]
    \centering
    \includegraphics[width=0.95\linewidth]{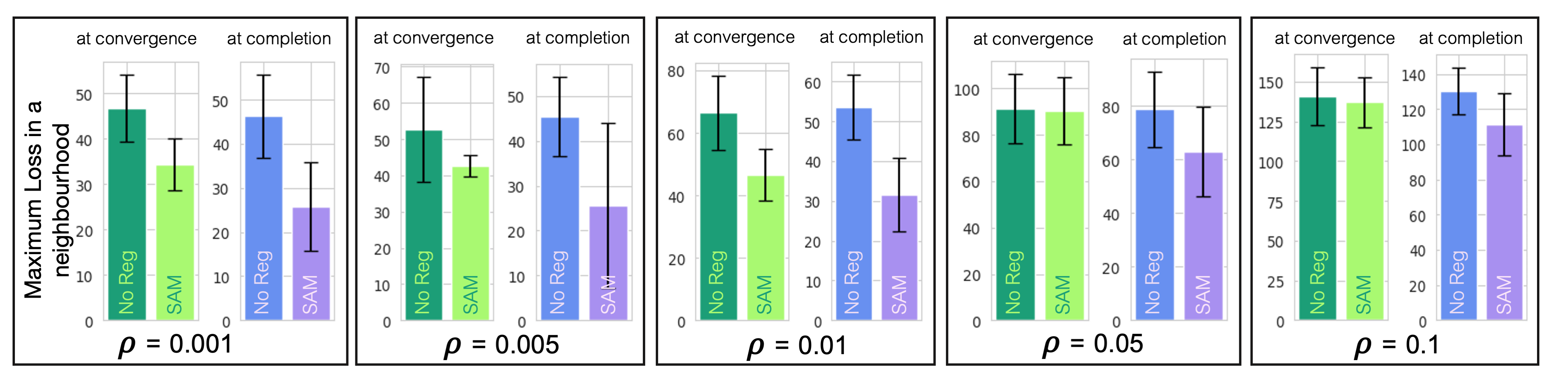}
    \caption{Meta-Test Testing Results for \textbf{CNN} on \textbf{FMNIST} dataset with \textbf{SAM} regularization.}
    \label{fig:conv_fmnist_sam}
\end{figure}

\begin{figure}[htbp]
    \centering
    \includegraphics[width=0.95\linewidth]{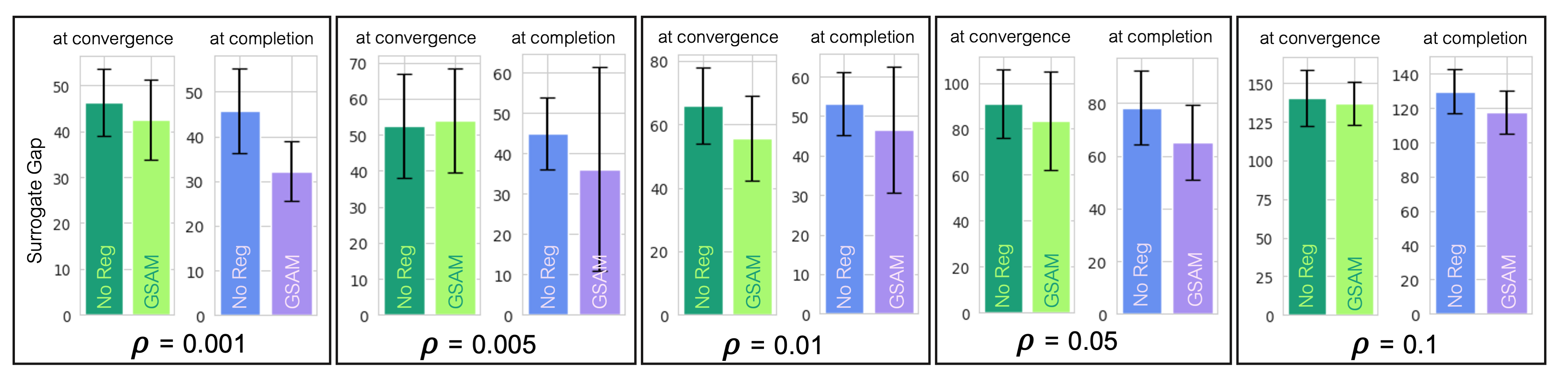}
    \caption{Meta-Test Testing Results for \textbf{CNN} on \textbf{FMNIST} dataset with \textbf{GSAM} regularization.}
    \label{fig:conv_fmnist_gsam}
\end{figure}

\begin{figure}[htbp]
    \centering
    \includegraphics[width=0.95\linewidth]{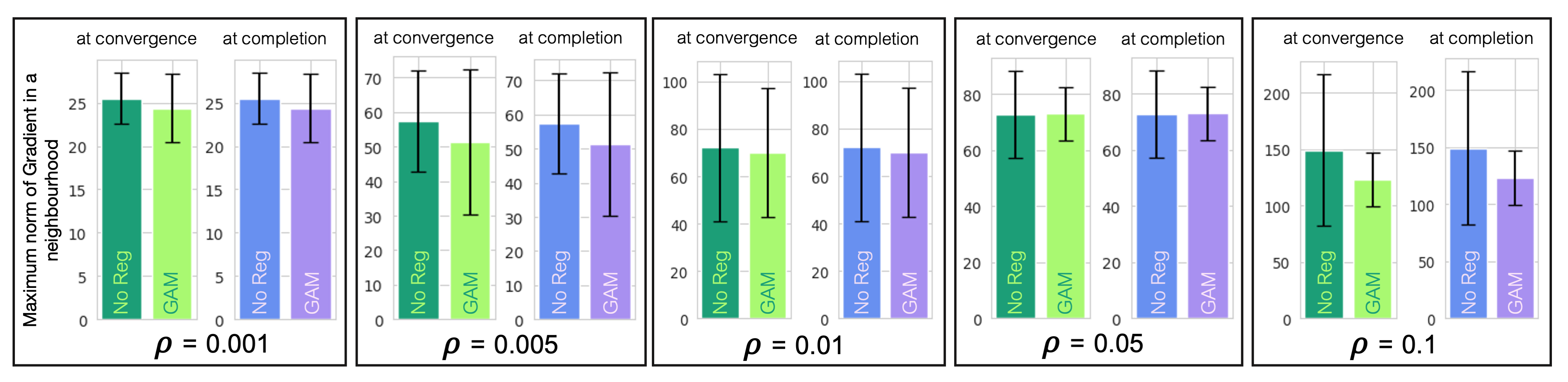}
    \caption{Meta-Test Testing Results for \textbf{CNN} on \textbf{FMNIST} dataset with \textbf{GAM} regularization.}
    \label{fig:conv_fmnist_gam}
\end{figure}

\textbf{TASK} : CIFAR10 Classification, using Convolution Neural Network (two 2D Convolution Layer with 16 3 × 3 kernels and 32 5 × 5 kernels with 2D Max Pool layer in between with kernel size 2 × 2)

\begin{figure}[htbp]
    \centering
    \includegraphics[width=0.95\linewidth]{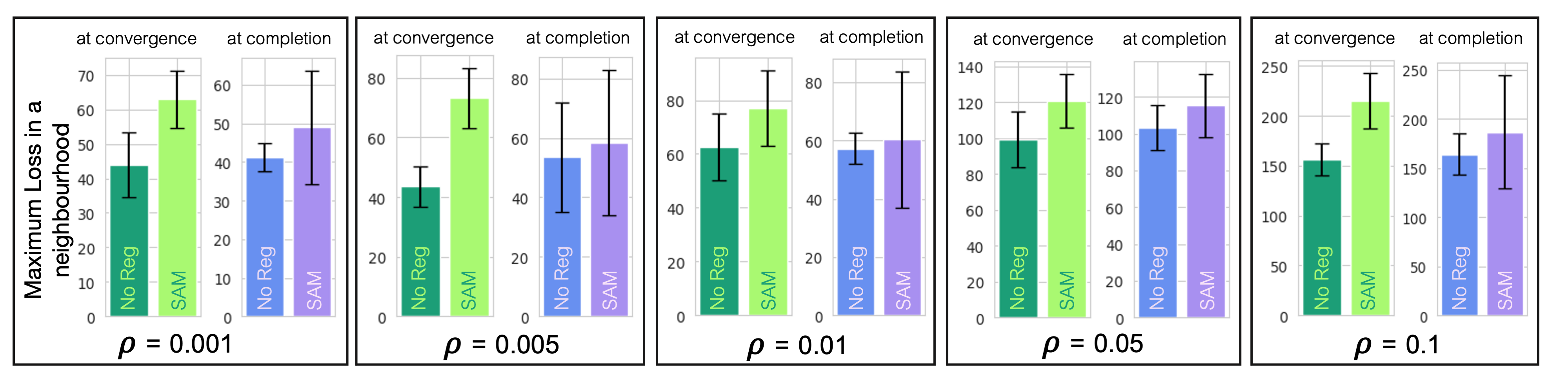}
    \caption{Meta-Test Testing Results for \textbf{CNN} on \textbf{CIFAR10} dataset with \textbf{SAM} regularization.}
    \label{fig:conv_cifar10_sam}
\end{figure}

\begin{figure}[htbp]
    \centering
    \includegraphics[width=0.95\linewidth]{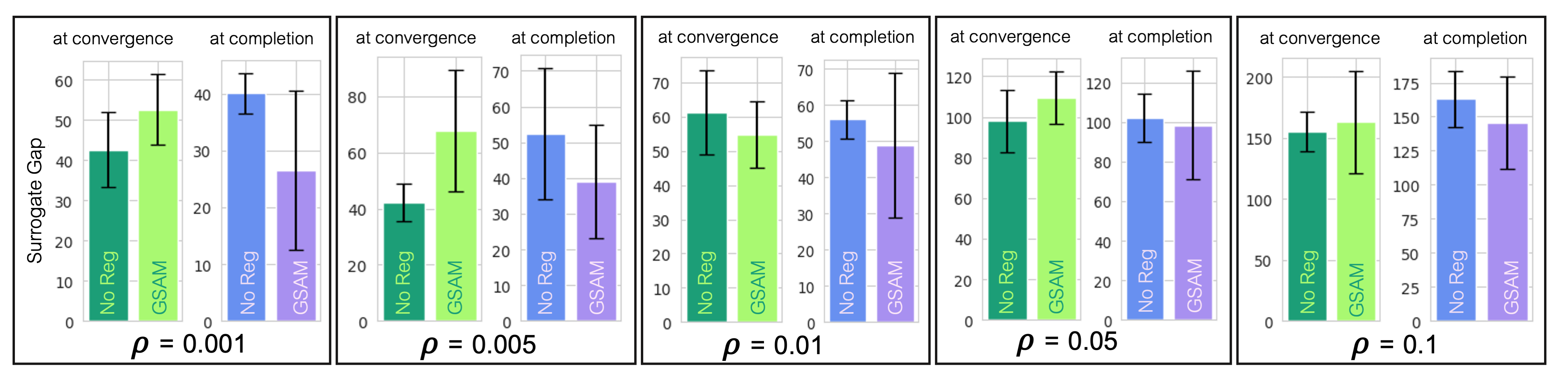}
    \caption{Meta-Test Testing Results for \textbf{CNN} on \textbf{CIFAR10} dataset with \textbf{GSAM} regularization.}
    \label{fig:conv_cifar10_gsam}
\end{figure}

\begin{figure}[htbp]
    \centering
    \includegraphics[width=0.95\linewidth]{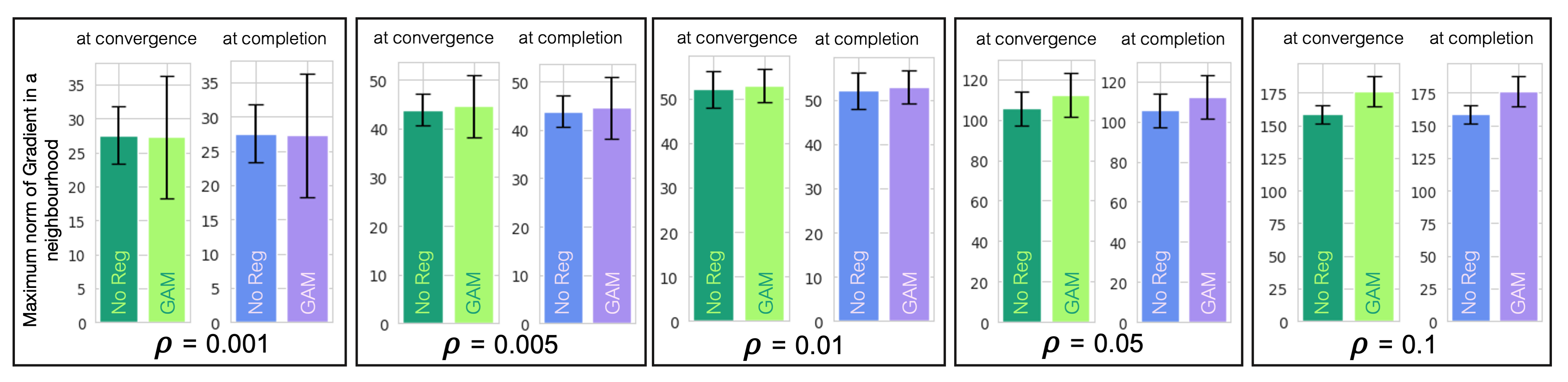}
    \caption{Meta-Test Testing Results for \textbf{CNN} on \textbf{CIFAR10} dataset with \textbf{GAM} regularization.}
    \label{fig:conv_cifar10_gam}
\end{figure}

\end{document}